%% file: emnlp2023.tex
\pdfoutput=1
\documentclass[11pt]{article}

\usepackage{EMNLP2023}

\usepackage{times}
\usepackage{latexsym}
\usepackage{amsmath}
\usepackage[most]{tcolorbox}

\usepackage[T1]{fontenc}
\usepackage[utf8]{inputenc}

\usepackage{microtype}

\usepackage{inconsolata}

\usepackage{multirow}
\usepackage{booktabs}
\usepackage{graphicx}
\usepackage{makecell}
\usepackage{hyperref}
\usepackage{tcolorbox}
\usepackage{booktabs,pifont}

\title{\texttt{TQLite}: Multi-LLM Jury Guided Distillation for Real-time MQM Translation Quality Evaluation}

\author{
Bhavin Jawade \and Cameron R. Wolfe \\
Netflix \\
\texttt{\{bjawade,cameronwolfe\}@netflix.com}
}

\begin{document}
\maketitle
\begin{abstract}
Large language models (LLMs) have demonstrated impressive performance in MQM-based translation quality (TQ) evaluation, and recent advances in large reasoning models (LRMs) promise even greater improvements.
However, both LLMs and LRMs are computationally expensive to deploy at scale, while small language models (SLMs)---\emph{though much more efficient}---struggle with the complex reasoning required for evaluation tasks.
In this work, we present an extensive empirical study benchmarking SLMs, LLMs, and LRMs across a wide range of TQ evaluation setups, providing a comprehensive view of the current landscape and establishing best practices.
To address the scalability challenge, we introduce \texttt{TQLite}, a novel distillation framework that enables SLMs to approach the MQM evaluation performance of the best LRM-based evaluators.
Our approach leverages a multi-LRM jury to generate high-quality synthetic training data via practical data curation techniques and aggregation of evaluation responses across a diverse panel of models.
Our results demonstrate that SLMs trained via \texttt{TQLite} achieve strong MQM evaluation performance that far exceeds off-the-shelf evaluation capabilities of standard SLMs, offering a scalable and cost-effective alternative to LLM- and LRM-based evaluators.
\end{abstract}

\section{Introduction}
% setting LLM context
Recent advancements in large language models (LLMs) have made them capable of accurately solving a wide variety of tasks \cite{brown2020language}.
However, the broad scope of LLMs makes them more difficult to evaluate compared to prior generations of deep neural networks.
Not only can LLMs solve many tasks, but the tasks that they solve tend to be open-ended---\emph{there may be many ``correct'' answers to any given question}.
In this regime, traditional evaluation metrics for text generation are less effective \cite{papineni2002bleu, banerjee2005meteor} and tend to correlate poorly with human judgments \cite{stiennon2020learning}, leading to the popularization of reference-free and open-ended evaluation techniques like LLM-as-a-Judge \cite{zheng2023judging}.

\begin{figure}
    \centering
    \includegraphics[width=1.0\linewidth]{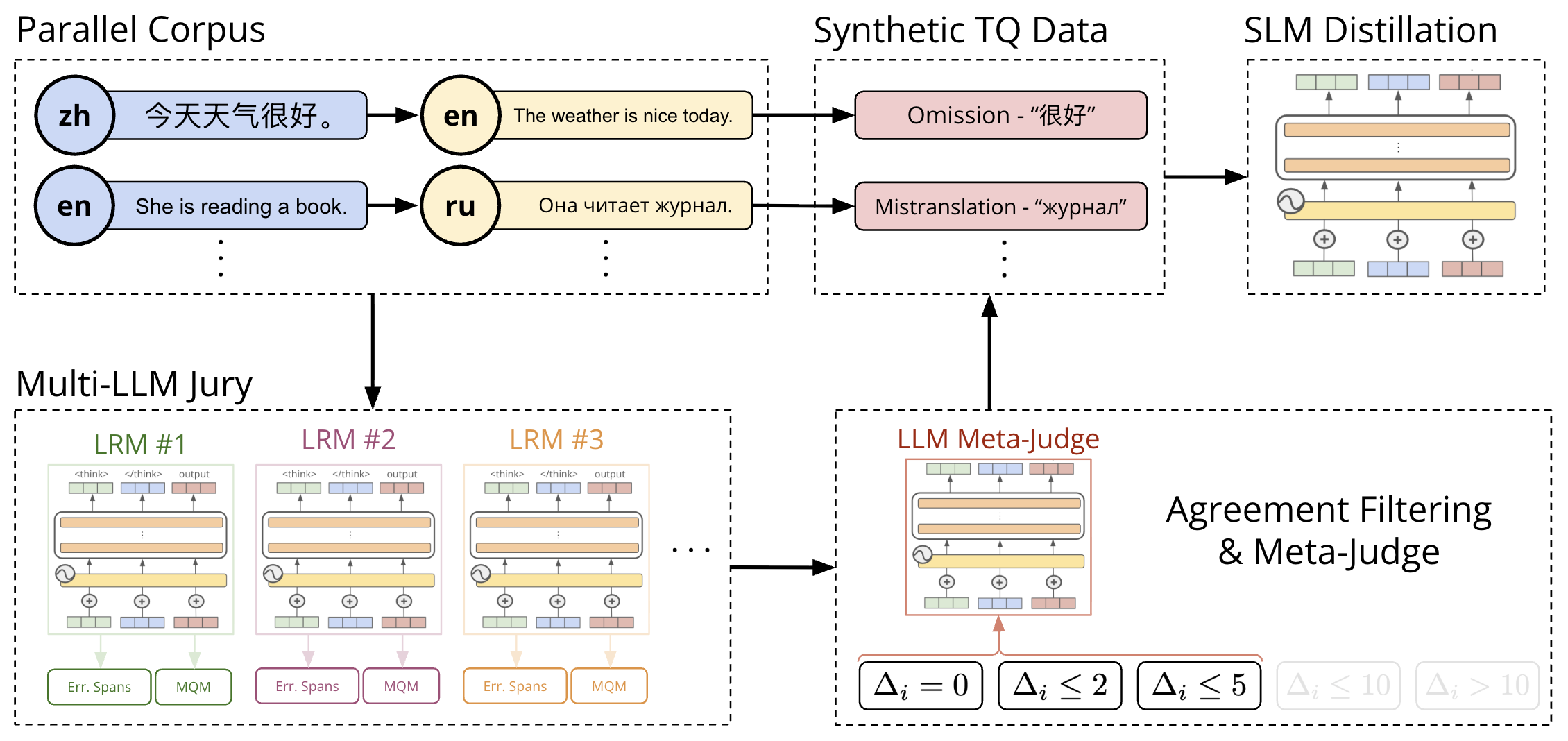}
    \caption{Our \texttt{TQLite} framework for distilling a multi-LRM jury into an SLM for efficient TQ evaluation.}
    \label{fig:tqlite_schematic}
\end{figure}

% llm-as-a-judge background
The concept of LLM-as-a-Judge is simple: \emph{we just prompt a powerful LLM to evaluate a segment of text}.
Usually, the structure and content of the prompt provided to the LLM judge resemble instructions that would normally be provided to a human evaluator \cite{chiang2023can}; e.g., a description of criteria to evaluate, a scoring rubric with detailed evaluation instructions, the scale or format of score to return, and more.
LLM-as-a-Judge has been shown to correlate well with human assessments of generation quality \cite{li2023alpacaeval}, leading to its widespread adoption as an evaluation technique for LLMs \cite{gu2024survey}.

% llm-as-a-judge for translation
LLM-as-a-Judge approaches have recently been adopted in research on machine translation (MT).
Prior learned metrics for evaluating translation quality (TQ) explored finetuning smaller language models over labels from human evaluators to detect translation errors at both a segment \cite{sellam2020bleurt, rei2022comet} and a span \cite{guerreiro2024xcomet} level.
Today, the best metrics for evaluating translation quality (TQ), as judged by performance on the WMT metrics task \cite{freitag2024llms}, are primarily LLM-based.
Several techniques have been proposed for applying LLM-as-a-Judge to TQ evaluation using both a direct assessment approach \cite{kocmi2023large} or a multidimensional quality metrics (MQM) framework \cite{kocmi2023gemba, fernandes2023devil}.

% where can we improve upon LLMs for TQ
When prompted with an MQM-based scoring framework, LLM judges provide state-of-the-art results for TQ evaluation at the system level \cite{kocmi2023gemba}.
At the segment level, the performance of such models naturally lags behind that of the best learned metrics \cite{juraska2024metricx}, but this gap can be closed with additional finetuning \cite{fernandes2023devil}. 
The possibility of evaluating TQ with smaller open-source models has been partially explored \cite{lu2024mqm}, but nearly all state-of-the-art LLM judges for TQ are based on large closed-source models (e.g., GPT-4 \cite{achiam2023gpt} or PaLM \cite{anil2023palm}), \emph{making the evaluation process both expensive and difficult to replicate}.

%  current research that impacts TQ
Despite this current dependence on closed-source models, these trends are likely to be disrupted in the dynamic environment of LLM research.
Given the close relationship between evaluation and reasoning, large reasoning models (LRMs)---\emph{including both closed \cite{jaech2024openai} and open-source \cite{guo2025deepseek, team2025kimi} variants}---have the potential to boost TQ evaluation capabilities, albeit at the cost of a larger computational footprint \cite{feng2025efficient}.
Additionally, higher-quality open-source LLMs are now available \cite{team2025gemma, yang2025qwen3} that may rival closed-source models in the context of TQ evaluation.

\vspace{0.1cm}
\noindent
\textbf{Our contribution.} We provide a comprehensive and practical benchmark of modern LLMs for TQ evaluation that covers models of all modes (i.e., reasoning versus standard), sizes, and levels of openness.
Such an analysis paints a clear picture of current LLM-as-a-Judge capabilities for TQ.
Based on the results of this analysis, we focus upon improving existing capabilities in three key areas:
\vspace{-0.1cm}
\begin{itemize}
    \item[1.] Cost.\vspace{-0.1cm}
    \item[2.] Performance.\vspace{-0.1cm}
    \item[3.] Openness.
\end{itemize}
\vspace{-0.1cm}
Frontier-level LRMs and LLMs already possess impressive evaluation capabilities, but we find that current open models---\emph{and especially small language models (SLMs)}---lag behind the TQ evaluation capabilities of top models.
To solve this issue, we develop ``TQLite'', a novel knowledge distillation framework that uses a jury of the top LRMs and LLMs to construct a high-quality synthetic training dataset for TQ evaluation via majority voting.
We show that TQLite can be used to create SLM-based judges for TQ evaluation that are open, efficient and competitive with state-of-the-art metrics at both the system and segment level.

\section{Related Work}

\textbf{Reasoning.}
Although early LLMs were criticized for their poor reasoning capabilities \cite{brown2020language}, later work on chain of thought (CoT) prompting found that LLMs can reason well when prompted to explain their output in either a few \cite{wei2022chain} or zero-shot \cite{kojima2022large} manner.
This discovery has led CoT prompting to become a staple for LLMs, and many variants have been proposed \cite{yu2023towards}.
Recent advancements in LRMs are related to work on CoT prompting, as LRMs operate by outputting a long CoT prior to producing their final output \cite{guo2025deepseek, team2025kimi}.
Instead of eliciting this behavior via prompting, however, LRMs are trained to leverage their long CoT---\emph{often called a ``reasoning trace'' or ``long CoT''}---using large-scale reinforcement learning with verifiable rewards \cite{lambert2024tulu, shao2024deepseekmath}.
The long CoT from an LRM is much longer than a standard CoT and can typically be controlled via a reasoning token to determine how much effort (i.e., low, medium or high) should be invested into solving a problem.
At higher reasoning efforts, the model will generate a longer CoT, thus investing more inference-time compute into a particular completion.

\vspace{0.1cm}
\noindent
\textbf{LLM-as-a-Judge.}
The idea of using LLMs for evaluation was popularized by \citeauthor{zheng2023judging}, which conducted a formal study of the LLM-as-a-Judge approach used in an earlier blog post from the same group \cite{vicuna2023vicuna}.
This formal proposal of LLM-as-a-Judge was followed by a swath of related research.
GPTScore \cite{fu2023gptscore} benchmarks many LLMs based on their ability to evaluate text generation tasks. 
AlpacaEval \cite{li2023alpacaeval} uses LLM judges to create an automated leaderboard for instruction-following LLMs.
G-Eval \cite{liu2023g} introduces an evaluation framework in which LLMs assess text quality using a detailed rubric and chain-of-thought (CoT) prompting.
\citeauthor{chiang2023can} and \citeauthor{wang2023chatgpt} explore the level of agreement between LLM-as-a-Judge and human evaluators.
\citeauthor{wang2023faireval} analyzes biases in LLM judges and proposes potential calibration methods.
Prometheus \cite{kim2023prometheus, kim2024prometheus} explores finetuning open-source models to match the evaluation capabilities of frontier-level LLMs. 

\vspace{0.1cm}
\noindent
\textbf{LLMs for TQ Evaluation.}
\texttt{Gemba} \cite{kocmi2023large} pioneered the use of LLMs for TQ evaluation and was later extended by \texttt{Gemba-MQM} \cite{kocmi2023gemba} to use an MQM-based evaluation framework.
EAPrompt \cite{lu2023error} aims to improve the segment-level performance of LLM evaluators for TQ by exploring several prompting strategies, including CoT prompting.
MQM-APE \cite{lu2024mqm} extends the \texttt{Gemba-MQM} framework by using LLM-generated post edits to identify the most relevant error spans.
AutoMQM \cite{fernandes2023devil} benchmarks various off-the-shelf and finetuned LLMs for TQ evaluation with an MQM framework. 

\vspace{0.1cm}
\noindent \textbf{SLMs and Distillation.}
LLM scaling laws indicate that LLM performance improves smoothly with scale \cite{kaplan2020scaling}.
However, standard Chinchilla-optimal scaling laws \cite{hoffmann2022training} only consider training costs, ignoring the long-term inference cost of a larger model.
High inference costs have increased interest in creating high-quality SLMs, which are usually developed either:
\begin{itemize}
    \item[1.] In parallel to a larger LLM---\emph{for the purpose of enabling faster, lower-cost experimentation}---as part of a multi-LLM model family; e.g., Llama-4-Scout \cite{meta2025llama4} or Gemma-3-4b-it \cite{team2025gemma}.
    \item[2.] By distilling the knowledge of a larger teacher model into a more efficient student model; e.g., DeepSeek-R1-Distill \cite{guo2025deepseek}.  
\end{itemize}
The topic of knowledge distillation \cite{hinton2015distilling} has been widely explored for both LLMs and SLMs \cite{xu2024survey}.

\section{Preliminaries}
\textbf{Data.} The majority of our experiments are conducted on the test set from the WMT22 metrics task \cite{freitag2022results}.
This test set is comprised of three language pairs: zh-en, en-de and en-ru. 
For each language pair, this dataset contains $\sim$2,000 sentences that are sampled from four possible domains: news, social, conversational and e-commerce.
Translations of these sentences are generated by 54 different human and machine translation systems, largely taken from the WMT22 general MT task \cite{kocmi2022findings}.
In total, the dataset contains 106,758 segments to be evaluated across the three language pairs. 
% TODO: add info on MT-Pref, add (empty) link to data we generate?

\vspace{0.1cm}
\noindent \textbf{MQM framework.}
As described in \cite{freitag2021experts}, we use human-labeled MQM scores \cite{lommel2014multidimensional}, which are available for the WMT22 metrics test set, as the gold standard for evaluating TQ.
In the MQM framework, humans are asked to mark individual error spans within a translation segment and label each error span with both a category and a severity (i.e., critical, major, or minor). 
Instead of asking annotators to directly provide a quality score, the final MQM score is automatically derived from these error spans.
Interestingly, many LLM-based TQ evaluators also adopt an MQM framework by asking the model to identify error spans and deriving a final score from these spans \cite{kocmi2023gemba, fernandes2023devil}.

\vspace{0.1cm}
\noindent \textbf{Metrics.}
We focus on two key metrics for measuring TQ evaluation performance:
\begin{itemize} \vspace{-0.2cm}
    \item System-level pairwise accuracy \cite{kocmi2021ship}.\vspace{-0.2cm}
    \item Segment-level pairwise accuracy with tie calibration, a.k.a. $\texttt{acc}^\star$ \cite{deutsch2023ties}.\vspace{-0.2cm}
\end{itemize}
We use the \href{https://github.com/google-research/mt-metrics-eval}{official scripts from WMT22}\footnote{We specifically adopt the evaluation code from the \href{https://github.com/MicrosoftTranslator/GEMBA}{\texttt{Gemba-MQM} repository} to ensure an exact match.} for evaluation and avoid reporting correlation-based metrics on both the system and segment level due to known sensitivities to small sample sizes and outliers \cite{mathur2020tangled}. 

\vspace{0.1cm}
\noindent\textbf{Baselines.}
We compare our TQ evaluation models to various baseline metrics and aim to cover a wide scope of methodologies in our analysis.
We consider reference-based learned metrics like MetricX \cite{juraska2024metricx}, COMET-22 \cite{rei2022comet}, and BLEURT \cite{sellam2020bleurt}. 
Additionally, learned reference-free metrics like COMET-kiwi \cite{rei2022cometkiwi} and COMET-QE \cite{rei2021references} are included, as well as LLM-based TQ evaluation metrics that precede \texttt{Gemba-MQM} like \texttt{Gemba-DA} \cite{kocmi2023large}. 

\begin{figure}
    \centering
    \includegraphics[width=1.0\linewidth]{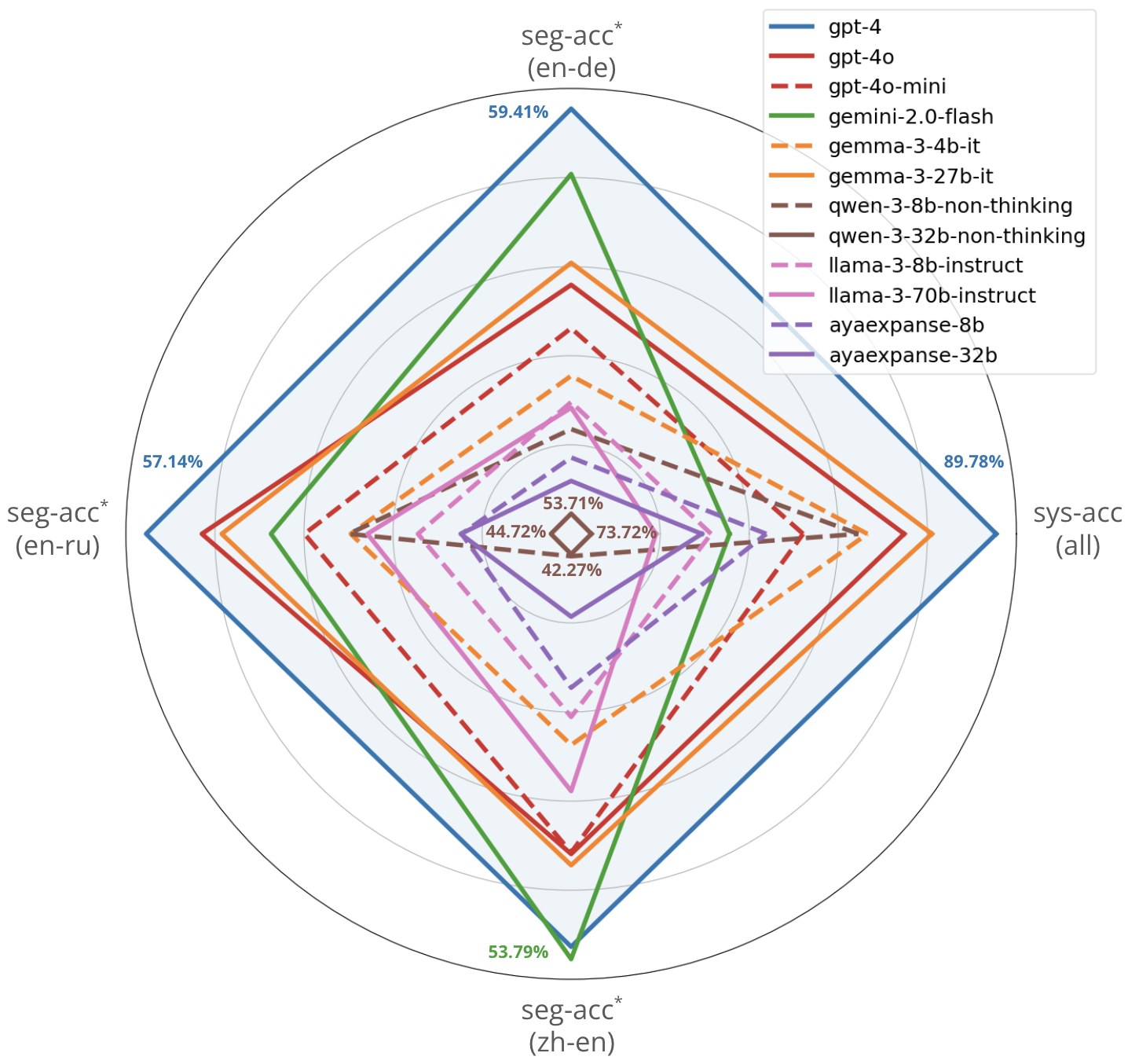}
    \caption{System-level (aggregate) and segment-level (per language pair) accuracy of LLMs using the standard \texttt{Gemba-MQM} prompt on the WMT22 metrics test set. Color indicates model family, while dashed lines differentiate between different members of the same model family. The axis for each metric is normalized using the minimum and maximum performance achieved by any LLM, and the figure is annotated with the best and worst performance along each axis. }
    \label{fig:gemba_mqm}
\end{figure}
\section{Benchmarking LLM Judges for TQ} \label{sec:bench_laaj}
Prior to training any SLM-based evaluators, we aim to develop a detailed understanding of the current limits and best practices for TQ evaluation with LLM-as-a-Judge.
We begin with the prompt and framework from \texttt{Gemba-MQM} \cite{kocmi2023gemba} and ablate various aspects of the evaluation setup while measuring performance on the WMT22 test set.
We match the settings of \citeauthor{kocmi2023gemba} by assuming a reference-free evaluation setting and keeping the same three few-shot examples fixed across all experiments.
Full experimental details are outlined in Appendix \ref{sec:apx:exp_details}, and all prompts used in our experiments are provided in Appendix \ref{sec:apx:prompts}.

\vspace{0.1cm}\noindent
\textbf{LLM selection.} % TODO: add all models
To begin, we replicate the prompt and scoring setting of \texttt{Gemba-MQM} across several closed and open-source LLMs, including GPT-4 \cite{achiam2023gpt}, GPT-4o and GPT-4o-mini \cite{hurst2024gpt}, Gemini-2.0-flash \cite{gemini2024},  Llama-3 \cite{dubey2024llama}, Aya-expanse \cite{dang2024aya}, Qwen-3 (non-thinking) \cite{yang2025qwen3} and Gemma-3 \cite{team2025gemma}.
The goal of this experiment is to $i)$ measure TQ evaluation capabilities across a wide set of different LLMs and $ii)$ arrive at a smaller representative set of LLMs to use for subsequent experiments. 

The results of this experiment are outlined in Figure \ref{fig:gemba_mqm}.
As we can see, the default \texttt{Gemba-MQM} prompt is specialized to the gpt-4 model---\emph{all other models, including the newer gpt-4o model, perform significantly worse than gpt-4 when using this prompt}.
The gemini-2.0-flash model performs well on a segment-level---\emph{even surpassing the performance of gpt-4 on zh-en evaluation}---but lags behind other models on a system level.
Additionally, we see that efficient model variants tend to be outperformed by their corresponding full models; e.g., gpt-4o-mini performs worse than gpt-4o.

When examining the performance of open LLMs, we see that the Gemma-3 family is a clear front-runner in TQ evaluation performance.
Although closed-source LLMs tend to achieve the best performance, Gemma-3-27b-it actually rivals or surpasses the performance of GPT-4o on all metrics. 
Interestingly, qwen-3 models are found to perform poorly on TQ evaluation.
We learn from manual inspection that this poor performance arises from the fact that qwen-3 models output a much larger number of error spans relative to other models.
This issue persisted after several attempted prompt modifications and despite adhering to recommended sampling settings.

Based on these results, we select the following LLMs for subsequent experiments: GPT-4o (closed LLM), GPT-4o-mini (closed SLM), Gemma-3-27b-it (open LLM), Gemma-3-4b-it (open SLM).
We omit GPT-4 despite its impressive performance because the model is older, more expensive, and subject to strict rate limits, which restricts experiments.

\begin{table*}[ht]
\footnotesize
\centering
\begin{tabular}{cclcccccc}
\toprule
\multirow{2}{*}{\textbf{Prompt}} & \multirow{2}{*}{\textbf{Output Format}} & \multirow{2}{*}{\textbf{Model}} &
\multicolumn{2}{c}{\textbf{System-Level Acc.}} &
\multicolumn{4}{c}{\textbf{Segment-Level Acc*}} \\
\cmidrule(lr){4-5} \cmidrule(lr){6-9}
& & &
\multicolumn{2}{c}{\textbf{All LPs}} &
\multicolumn{1}{c}{\textbf{En-De}} &
\multicolumn{1}{c}{\textbf{En-Ru}} &
\multicolumn{1}{c}{\textbf{Zh-En}} &
\multicolumn{1}{c}{\textbf{Avg.}} \\
\toprule
\multicolumn{2}{c}{\multirow{6}{*}{Baselines}} & \texttt{bleurt-20} & \multicolumn{2}{c}{84.67\%} & 56.89\% & 53.37\% & 49.83\% & 53.36\% \\
& & \texttt{comet-22} & \multicolumn{2}{c}{83.94\%} & 60.04\% & 58.34\% & 53.93\% & 57.44\% \\
& & \texttt{metricx-xxl-MQM-2020} & \multicolumn{2}{c}{85.04\%} & 61.51\% & 60.70\% & 54.29\% & 58.83\% \\
& & \texttt{comet-qe} & \multicolumn{2}{c}{78.10\%} & 56.96\% & 53.48\% & 49.05\% & 53.16\% \\
& & \texttt{comet-kiwi} & \multicolumn{2}{c}{78.83\%} & 57.67\% & 56.33\% & 51.77\% & 55.26\% \\
& & \texttt{gemba-gpt4-DA} & \multicolumn{2}{c}{87.59\%} & 56.20\% & 52.12\% & 53.76\% & 54.03\% \\

\midrule
\multirow{5}{*}{\texttt{Gemba-MQM}} & \multirow{5}{*}{Free Text} & \texttt{gpt-4} & \multicolumn{2}{c}{89.79\%} & 59.41\% & 57.14\% & 53.44\% & 56.66\% \\
& & \texttt{gpt-4o} & \multicolumn{2}{c}{86.13\%} & 56.93\% & 55.43\% & 50.80\% & 54.39\% \\
& & \texttt{gpt-4o-mini} & \multicolumn{2}{c}{82.12\%} & 57.19\% & 53.25\% & 50.23\% & 53.56\% \\
& & \texttt{gemma-3-4b-it} & \multicolumn{2}{c}{84.67\%} & 55.65\% & 50.86\% & 47.70\% & 51.40\% \\
& & \texttt{gemma-3-27b-it} & \multicolumn{2}{c}{87.23\%} & 57.24\% & 54.82\% & 51.12\% & 54.39\% \\

\midrule
\multirow{12}{*}{\texttt{TQLite}} & \multirow{4}{*}{Free Text} & \texttt{gpt-4o} &\multicolumn{2}{c}{85.87\%} & 57.03\% & 54.92\% & 50.34\% & 54.10\% \\
& & \texttt{gpt-4o-mini} & \multicolumn{2}{c}{78.10\%} & 54.62\% & 49.77\% & 48.39\% & 50.92\% \\
&& \texttt{gemma-3-4b-it} & \multicolumn{2}{c}{81.39\%} & 55.96\% & 49.50\% & 45.97\% & 50.48\% \\
&& \texttt{gemma-3-27b-it} & \multicolumn{2}{c}{86.86\%} & 56.29\% & 52.63\% & 50.93\% & 52.28\% \\

\cmidrule(lr){2-9}& \multirow{4}{*}{\shortstack{Structured Text}} & \texttt{gpt-4o} & \multicolumn{2}{c}{84.67\%} & 56.79\% & 53.50\% & 49.21\% & 53.17\% \\
& & \texttt{gpt-4o-mini} & \multicolumn{2}{c}{78.10\%} & 54.87\% & 50.50\% & 47.40\% & 50.93\% \\
&& \texttt{gemma-3-4b-it} & \multicolumn{2}{c}{77.74\%} & 54.91\% & 49.32\% & 42.39\% & 48.87\% \\
&& \texttt{gemma-3-27b-it} & \multicolumn{2}{c}{84.31\%} & 55.91\% & 53.15\% & 50.42\% & 53.16\% \\

\cmidrule(lr){2-9} & \multirow{4}{*}{JSON}& \texttt{gpt-4o} & \multicolumn{2}{c}{82.48\%} & 56.11\% & 52.00\% & 48.15\% & 52.09\% \\
& & \texttt{gpt-4o-mini} & \multicolumn{2}{c}{73.72\%} & 55.47\% & 52.26\% & 49.37\% & 52.36\% \\
&& \texttt{gemma-3-4b-it} & \multicolumn{2}{c}{77.37\%} & 54.48\% & 46.78\% & 44.09\% & 48.45\% \\
&& \texttt{gemma-3-27b-it} & \multicolumn{2}{c}{85.04\%} & 56.58\% & 53.12\% & 51.23\% & 53.64\% \\

\midrule
\multirow{8}{*}{\texttt{TQLite-ST}} & \multirow{4}{*}{\shortstack{Structured Text}} & \texttt{gpt-4o} & \multicolumn{2}{c}{84.31\%} & 57.63\% & 54.81\% & 50.27\% & 54.23\% \\
& & \texttt{gpt-4o-mini} & \multicolumn{2}{c}{82.48\%} & 57.56\% & 54.91\% & 47.80\% & 53.42\% \\
&& \texttt{gemma-3-4b-it} & \multicolumn{2}{c}{76.28\%} & 55.72\% & 50.27\% & 44.07\% & 50.02\% \\
&& \texttt{gemma-3-27b-it} & \multicolumn{2}{c}{84.67\%} & 57.28\% & 54.77\% & 51.06\% & 54.37\% \\

\cmidrule(lr){2-9} & \multirow{4}{*}{\shortstack{CoT +\\Structured Text}} & \texttt{gpt-4o} & \multicolumn{2}{c}{83.21\%} & 58.16\% & 55.83\% & 51.96\% & 55.32\% \\
& & \texttt{gpt-4o-mini} & \multicolumn{2}{c}{75.55\%} & 56.74\% & 53.21\% & 47.55\% & 52.50\% \\
&&\texttt{gemma-3-4b-it} & \multicolumn{2}{c}{80.66\%} & 55.24\% & 49.22\% & 45.83\% & 50.10\% \\
&& \texttt{gemma-3-27b-it} & \multicolumn{2}{c}{86.50\%} & 56.29\% & 52.95\% & 50.24\% & 53.16\% \\

% \cmidrule(lr){2-9} & \multirow{8}{*}{\shortstack{Long CoT +\\Structured Text}} &\texttt{qwen3-8b-thinking} & \multicolumn{2}{c}{76.64\%} & 54.10\% & 48.27\% & 48.79\% & 50.39\% \\
% &&\texttt{qwen3-32b-thinking} & \multicolumn{2}{c}{83.58\%} & 55.46\% & 52.54\% & 49.76\% & 52.59\% \\
% && \texttt{o1-low} & \multicolumn{2}{c}{85.04\%} & 58.88\% & 57.05\% & 53.75\% & 56.56\% \\
% && \texttt{o1-high} & \multicolumn{2}{c}{86.13\%} & 57.71\% & 57.62\% & 53.79\% & 56.37\% \\
% && \texttt{o3-low} & \multicolumn{2}{c}{91.61\%} & 58.08\% & 58.63\% & 51.92\% & 56.21\% \\
% && \texttt{o3-high} & \multicolumn{2}{c}{92.34\%} & 57.99\% & 57.49\% & 52.88\% & 56.12\% \\
% && \texttt{o4-mini-low} & \multicolumn{2}{c}{89.05\%} & 57.45\% & 55.94\% & 50.92\% & 54.77\% \\
% && \texttt{o4-mini-high} & \multicolumn{2}{c}{87.96\%} & 56.35\% & 55.47\% & 50.19\% & 54.00\% \\

\bottomrule
\end{tabular}
\caption{System and segment-level accuracy of various prompt and model combinations on the WMT22 test set.}
\label{tab:llm_baselines}
\end{table*}

\vspace{0.1cm}
\noindent \textbf{Prompt modifications.}
The original \texttt{Gemba-MQM} prompt has some sources of ambiguity that would be beneficial to resolve.
For example, the prompt does not explicitly specify the expected format of the model's output and uses the word ``categories'' to describe both error categories and severities.
To solve these issues, we create a slightly modified prompt for TQLite that resolves this ambiguity by providing a separate, well-structured description of error categories and severity levels; see Appendix \ref{sec:apx:prompts} for the full prompt.

Given the impressive performance of the \texttt{Gemba-MQM} prompt, we maintain the core information from this prompt and keep changes to a minimum while solving the aforementioned issues.
As shown in Table \ref{tab:llm_baselines}, the TQLite prompt with free text output still performs slightly worse than the original Gemba-MQM prompt, indicating that performance on the WMT22 metrics test set is sensitive to small changes.
However, we choose to avoid improving performance via manual prompt optimizations.
The experiments in this section focus upon measuring the relative performance impact of systematic changes to the TQ evaluation setup, making absolute performance less of a concern.
Our core TQLite experiments are also primarily based on finetuning, where prompt engineering has less of an impact.

\vspace{0.1cm}
\noindent \textbf{Structured outputs.}
The \texttt{Gemba-MQM} prompt provides no explicit description of the model's expected output structure, instead choosing to demonstrate the correct output structure via few-shot examples.
As a result, adherence to the desired output structure is noisy. 
For example, $\sim$10\% of \texttt{Gemba-MQM} outputs with GPT-4 contain a partial formatting error\footnote{Partial formatting errors occur when the model does not cover all three error severities in its output or provides an error without a properly formatted span.}, and this ratio increases to 85\% for less powerful models like Gemma-3-4B-it.

For this reason, parsing the evaluation results from \texttt{Gemba-MQM} and \texttt{TQLite} requires the use of many custom---\emph{and imperfect}---heuristics that, if implemented incorrectly, can drastically influence evaluation results.
To prevent problems caused by heuristic parsing, we can give the model detailed instructions on how to format its output.
However, such format restrictions have the potential to deteriorate the LLM's performance \cite{tam2024let}.

In Table \ref{tab:llm_baselines}, we test two variants of the \texttt{TQLite} prompt that instruct the model to format its output either as structured text \footnote{The structured text approach matches the original output format of \texttt{Gemba-MQM}, but we add explicit instructions describing this structure into the prompt.} or a JSON document.
Enforcing a strict output structure degrades TQ evaluation quality across all models at a system level, and JSON formatting seems to be slightly more detrimental than structured text formatting.
Notably, the largest degradation in TQ evaluation quality occurs in the less powerful SLM models, \emph{indicating that models with less capacity struggle to handle more complex evaluation instructions}.

Despite these results, the need for heuristic parsing can significantly complicate the process of deriving the final MQM score from a model’s predicted output.
In our experiments, we observed that even minor adjustments to the parsing algorithm could substantially improve or degrade evaluation metrics, even when the model’s output is kept fixed.
Ideally, our TQ evaluator should adhere to a stricter output format to minimize external factors---\emph{unrelated to the quality of the evaluation model}---that could influence results.
For this reason, we adopt the structured text output format (i.e., the less problematic of the two options) for the remainder of our experiments, and we aim to develop models capable of following this format while maintaining high TQ evaluation quality.

\vspace{0.1cm}
\noindent \textbf{Multi vs. Single-Turn Prompts.}
The \texttt{Gemba-MQM} prompt uses a multi-turn chat format for its few-shot examples.
The evaluation instructions are presented to the model as a user message in each turn, and the desired output is provided as a response from the assistant. 
The segment to be scored is provided to the model as a final user message, and the model is expected to output the corresponding evaluation result.
Alternatively, we could derive a single-turn format for this prompt that includes all evaluation instructions and few-shot examples in a single user message. 
Such an approach simplifies the prompt and reduces the total number of tokens (i.e., by not repeating instructions in each turn) consumed by the model.

We create a single-turn variant of the \texttt{TQLite} prompt, which we call \texttt{TQLite-ST}; see Appendix \ref{sec:apx:prompts} for the full prompt.
We see in Table \ref{tab:llm_baselines} that using a single-turn chat format does not meaningfully impact TQ evaluation quality for any model.
In fact, \emph{both GPT-4o and Gemma-3-27b-it slightly improve at the segment level by using a single-turn prompt}, leading us to adopt single-turn prompts in the remainder of our experiments.
We also extend this single-turn prompt by asking the model to output an explanation along with its error spans (i.e., zero-shot CoT prompting \cite{kojima2022large}) but find that this approach---\emph{though moderately helpful in some cases}---does not yield a consistent or significant performance boost.

\begin{table*}[ht]
\footnotesize
\centering
\begin{tabular}{clcccccc}
\toprule
\multirow{2}{*}{\textbf{Category}} & \multirow{2}{*}{\textbf{Model}} &
\multicolumn{2}{c}{\textbf{System-Level Acc.}} &
\multicolumn{4}{c}{\textbf{Segment-Level Acc*}} \\
\cmidrule(lr){3-4} \cmidrule(lr){5-8}
& &
\multicolumn{2}{c}{\textbf{All LPs}} &
\multicolumn{1}{c}{\textbf{En-De}} &
\multicolumn{1}{c}{\textbf{En-Ru}} &
\multicolumn{1}{c}{\textbf{Zh-En}} &
\multicolumn{1}{c}{\textbf{Avg.}} \\
\toprule
\multirow{4}{*}{\shortstack{Open LRM}} & \texttt{qwen-3-8b-thinking} & \multicolumn{2}{c}{76.64\%} & 54.20\% & 48.44\% & 49.07\% & 50.57\% \\
&\texttt{qwen-3-32b-thinking} & \multicolumn{2}{c}{83.58\%} & 55.61\% & 51.78\% & 50.83\% & 52.74\% \\
&\texttt{deepseek-r1-distill-llama-8b} & \multicolumn{2}{c}{76.28\%} & 54.15\% & 45.47\% & 45.61\% & 48.41\% \\
&\texttt{deepseek-r1-distill-llama-70b} & \multicolumn{2}{c}{84.31\%} & 55.10\% & 51.10\% & 48.98\% & 51.73\% \\
\cmidrule(lr){1-8} \multirow{6}{*}{\shortstack{Closed LRM}} & \texttt{o1-low} & \multicolumn{2}{c}{85.04\%} & 58.88\% & 57.05\% & 53.75\% & 56.56\% \\
& \texttt{o1-high} & \multicolumn{2}{c}{86.13\%} & 57.71\% & 57.62\% & 53.79\% & 56.37\% \\
& \texttt{o3-low} & \multicolumn{2}{c}{91.61\%} & 58.08\% & 58.63\% & 51.92\% & 56.21\% \\
& \texttt{o3-high} & \multicolumn{2}{c}{92.34\%} & 57.99\% & 57.49\% & 52.88\% & 56.12\% \\
& \texttt{o4-mini-low} & \multicolumn{2}{c}{89.05\%} & 57.45\% & 55.94\% & 50.92\% & 54.77\% \\
& \texttt{o4-mini-high} & \multicolumn{2}{c}{87.96\%} & 56.35\% & 55.47\% & 50.19\% & 54.00\% \\
& \texttt{gemini-2.5-flash-thinking} & \multicolumn{2}{c}{87.23\%} & 58.17\% & 57.12\% & 52.24\% & 55.84\% \\
& \texttt{gemini-2.5-pro-thinking} & \multicolumn{2}{c}{89.42\%} & 58.90\% & 58.18\% & 54.12\% & 57.06\% \\

\cmidrule(lr){1-8} \multirow{2}{*}{\shortstack{Multi-LRM}} & \texttt{o4-mini-high} $\times 3$ & \multicolumn{2}{c}{87.96\%} & 59.03\% & 57.35\% & 52.48\% & 56.29\% \\
& \texttt{o4-mini-high} $\times$ \texttt{o3-high} $\times$ \texttt{o1-high} & \multicolumn{2}{c}{89.78\%} & 58.84\% & 58.74\% & 53.64\% & 57.07\% \\
\bottomrule
\end{tabular}
\caption{Performance of LRMs on the WMT22 test set using the \texttt{TQLite-ST} prompt with structured text outputs. }
\label{tab:lrm-tqlite-table}
\end{table*}

\vspace{0.1cm}
\noindent \textbf{Reasoning models.} Given that evaluation is fundamentally a reasoning problem, we conclude this initial LLM-as-a-Judge exploration by exploring the use of LRMs for TQ evaluation.
We consider a wide scope of open and closed-source LRMs, including deepseek-r1-distill-llama \cite{guo2025deepseek}, qwen-3 (thinking) \cite{yang2025qwen3}, gemini-2.5 \cite{comanici2025gemini}, o1 \cite{jaech2024openai}, o3 and o4-mini \cite{openai2025o3}.
For LRMs with reasoning token support, we test both low and high reasoning efforts.
All tests use the \texttt{TQLite-ST} prompt, and results are provided in Table \ref{tab:lrm-tqlite-table}.

LRMs achieve (by far) the best performance on the system level of any model considered so far.
In particular, o3 with high reasoning effort achieves a state-of-the-art system-level accuracy of 92.34\%.
At the segment level, TQ evaluation results with LRMs are more skewed.
For example, gemini-2.5-pro-thinking achieves an average segment-level accuracy of 57.06\%, outperforming all other LLM or LRM-based evaluators and approaching the performance of top learned metrics.
The performance of both o1 and o3 falls within a similar range, but other LRMs perform noticeably worse.

Open-source LRMs perform noticeably worse than their closed-source counterparts.
We observe that model size tends to benefit evaluation performance for both open and closed-source models; e.g., gemini-2.5-pro outperforms gemini-2.5-flash, while qwen-3-32b outperforms qwen-3-8b.
Additionally, increasing the reasoning effort improves performance in most cases, but o4-mini is found to perform slightly better with low reasoning effort. 

\begin{figure}
    \centering
    \includegraphics[width=\linewidth]{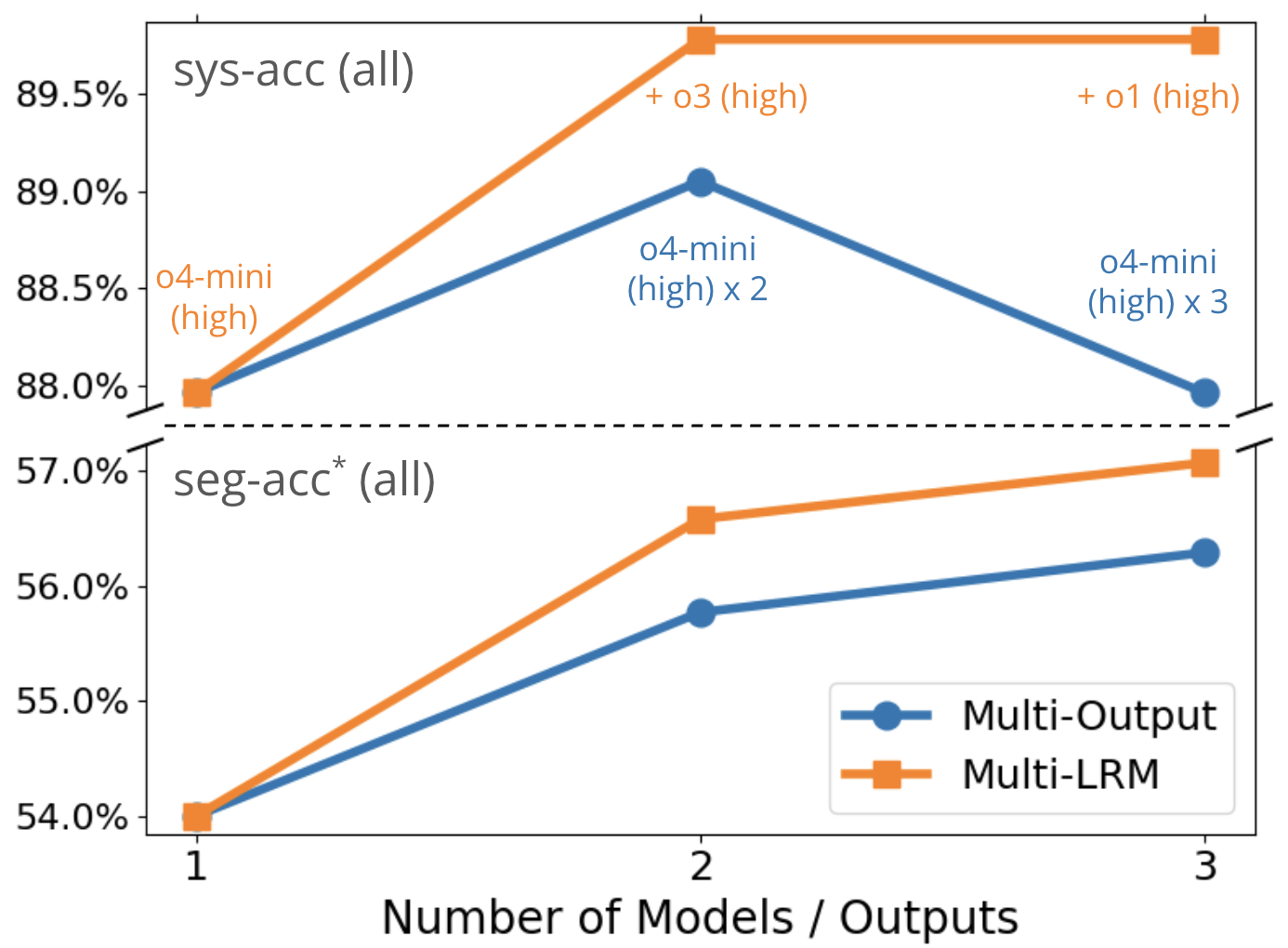}
    \caption{Aggregate system and segment-level accuracy on the WMT22 metrics test set as we average segment-level MQM scores from the same LRM (blue) or different LRMs (orange).}
    \label{fig:multi_lrm}
\end{figure}

\vspace{0.1cm}
\noindent \textbf{Multi-LRM jury.}
To further optimize performance, we can generate multiple LRM outputs for each segment and aggregate by computing the average score on a segment level.
This approach is common for LRMs and is even explicitly recommended in the \href{https://huggingface.co/deepseek-ai/DeepSeek-R1}{usage guidelines} for deepseek-r1.
We test two variants of a multi-LRM jury:
\begin{itemize}\itemsep0em
    \item \emph{Multi-Output}: generate multiple outputs from the same LRM. \vspace{-0.1cm}
    \item \emph{Multi-LRM}: generate multiple outputs from several different LRMs. \vspace{-0.1cm}
\end{itemize}
In both cases, we obtain our final score by taking the average segment-level score. 
This multi-LRM jury strategy is tested with o4-mini in Table \ref{tab:lrm-tqlite-table}, where we see that both aggregation strategies improve TQ evaluation performance.
Interestingly, this approach has the biggest impact at a segment level---\emph{using a multi-LRM jury yields a segment-level accuracy exceeding that of all individual LRMs in the jury}.

The behavior of multi-LRM aggregation is illustrated more granularly in Figure \ref{fig:multi_lrm}.
Beginning with o4-mini, this figure displays the change in system and segment-level accuracy as we increase the number of LRM outputs over which aggregation is performed.
As we can see, \emph{aggregating over outputs from different LRMs yields a more consistent performance improvement compared to aggregating over several outputs from the same LRM}.

\vspace{0.1cm}
\noindent \textbf{Takeaways.}
The findings of our LLM-as-a-Judge benchmark for TQ can be summarized as follows:
\begin{itemize}\itemsep0em \vspace{-0.2cm}
    \item Accurate TQ evaluation can be performed with both closed and open-source LLMs. \vspace{-0.2cm}
    \item SLMs perform poorly, especially when asked to strictly format their output. \vspace{-0.2cm}
    \item LRMs achieve the best performance and can be enhanced via a multi-LRM jury. \vspace{-0.2cm} 
    \item Closed-source models outperform open-source models in nearly all cases. \vspace{-0.2cm}
\end{itemize}
Based on these findings, we now present our \texttt{TQLite} methodology, which aims to distill the impressive TQ evaluation capabilities of (multi-)LRM judges into small open-source models.

\section{Distilling Translation Quality}
\label{sec:distill}

\paragraph{Problem setup.}
Let $\mathcal{D}=\{x_i\}_{i=1}^N$ denote a set of source-target examples, where
\[
x_i=\big(s_l^{(i)},\, s_t^{(i)},\, t_t^{(i)},\, t_l^{(i)}\big),
\]
with $s_l^{(i)}$ and $t_l^{(i)}$ the source/target language identifiers, $s_t^{(i)}$ the source text, and $t_t^{(i)}$ the translation.
We adopt an MQM-style schema $\Sigma=(\mathcal{C},\mathcal{V})$ with categories $\mathcal{C}$ and severities $\mathcal{V}$.
An error span annotation (ESA) is
\[
a=\big(e,\, c,\, v\big),
\]
with $e$ being the source or target error span, $c\in\mathcal{C}$ error category, $v\in\mathcal{V}$ the error severity. 

\vspace{0.1cm}
\noindent \textbf{Synthetic data generation.} We generate ESAs for each $x_i \in \mathcal{D}$ using a panel of $M$ large reasoning models (LRMs), denoted
\[
\mathcal{T} = \{f_{\theta_1}, \dots, f_{\theta_M}\}.
\]
Given the same prompt $P$ and schema $\Sigma$, each $f_{\theta_m}$ produces a set of ESAs $\mathcal{A}_i^{(m)}$ and a corresponding MQM score $S_i^{(m)}$. The MQM score for a set of ESAs is computed as the weighted sum of severity levels, normalized by the length of the translation segment.

To assess the level of agreement between the LRMs for a given segment, we compute the \emph{score range} as follows:
\[
\Delta_i = \max_{m} S_i^{(m)} - \min_{m} S_i^{(m)},
\]
which measures the spread between the most lenient and most critical judgments across the jury. Using this range, we categorize each example into one of five agreement levels:
\begin{itemize}\itemsep0em
    \item Perfect agreement: $\Delta_i = 0$
    \item High agreement: $\Delta_i \leq 2$
    \item Moderate agreement: $\Delta_i \leq 5$
    \item Low agreement: $\Delta_i \leq 10$
    \item No agreement: $\Delta_i > 10$
\end{itemize}

\input{./tables/tqlite_distilled_table.tex}

This categorization allows us to analyze where the jury of LRMs is consistent and where it diverges. Segments with high or perfect agreement are strong candidates for direct inclusion in the synthetic training set.

For each segment in the \textit{perfect} or \textit{high agreement} categories, we perform an additional refinement step using a \emph{meta-judge} LLM. The input to the meta-judge consists of the error spans produced by all LRMs for that segment, along with their corresponding error explanations. The meta-judge is prompted to:
\begin{enumerate}\itemsep0em
    \item Identify cases where different LRMs have highlighted spans that refer to the same underlying translation error, even if the exact span boundaries differ.
    \item Compare the associated error explanations to confirm that the described errors are semantically equivalent.
    \item Merge such cases into a single, aggregated error annotation, selecting canonical span boundaries and assigning the most appropriate category and severity.
\end{enumerate}

This process serves two purposes: (i) to remove duplicate annotations that arise from minor boundary or phrasing differences across LRMs, and (ii) to ensure consistency in the assigned categories and severities. The result is a clean, consolidated set of error span annotations for high-confidence segments, which is then used as synthetic supervision for training the distilled SLM evaluator.

\vspace{0.1cm}
\noindent \textbf{Finetuning the student model.} Using the filtered and cleaned high-quality annotations obtained from above, we train a smaller open-source language model to replicate the translation quality evaluation capabilities of the LRM jury. In our experiments, we use instruction-tuned models such as \texttt{Gemma-12B-it} and \texttt{Gemma-27B-it} as the student models. These models are \emph{orders of magnitude smaller} than typical LRMs used in the jury, which can exceed $10^{11}$–$10^{12}$ parameters.

We adopt a standard supervised fine-tuning (SFT) approach. Each training example consists of the same prompt format used during the data curation stage—this includes the original segment (source and translation), the evaluation instructions, and the expected structured output containing the error span annotations. By keeping the prompt format identical, we ensure that the student model learns the task under the same input–output specification as the LRMs from which the annotations were distilled.

\begin{figure}
    \centering
    \includegraphics[width=\linewidth]{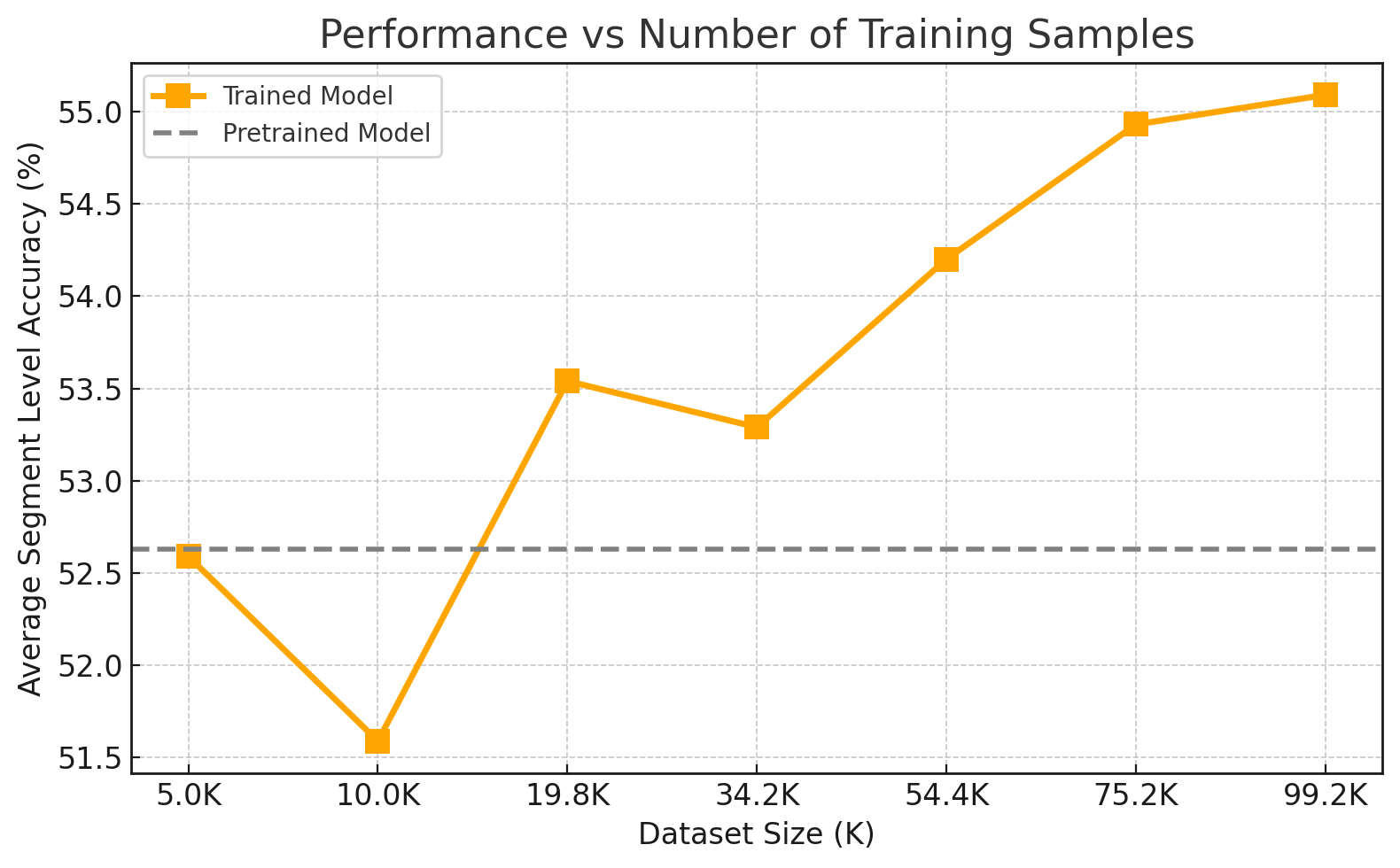}
    \caption{\textbf{Distillation Scaling -} Change in average segment-level accuracy with increase in number of training samples for \texttt{Gemma-3-12b-it}.}
    \label{fig:distillation_scaling}
\end{figure}

Formally, given a training set $\{(p_j, y_j)\}_{j=1}^K$, where $p_j$ is the prompt and $y_j$ is the meta-judge’s aggregated output, we minimize the standard negative log-likelihood:
\[
\mathcal{L}_{\mathrm{SFT}} = -\frac{1}{K} \sum_{j=1}^K \log P_{\psi}(y_j \mid p_j),
\]
where $\psi$ are the parameters of the student model. We train until convergence on a held-out synthetic validation set.

\begin{figure}
    \centering
    \includegraphics[width=\linewidth]{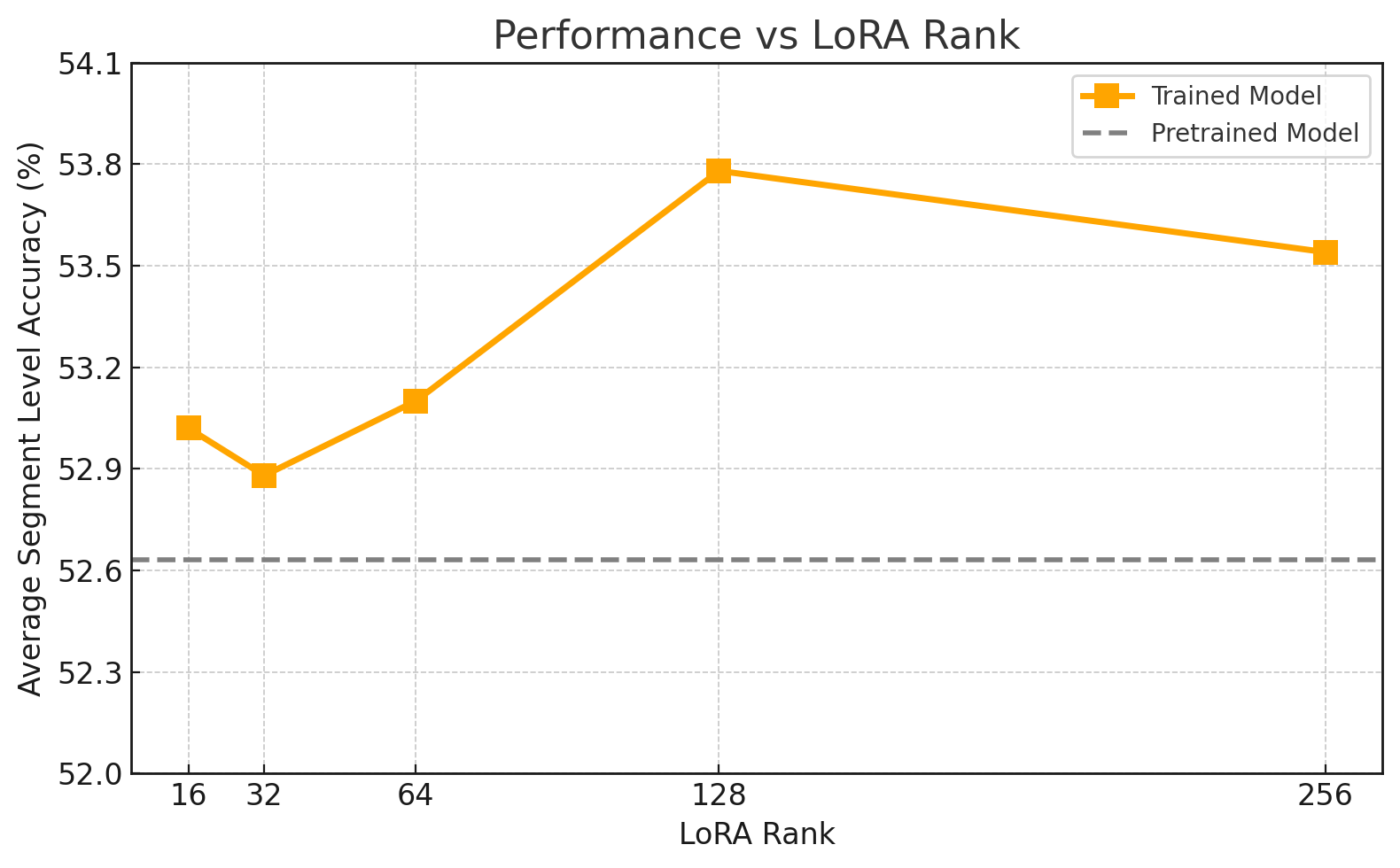}
    \caption{Impact of LoRA Rank (number of trainable parameters) on average segment-level accuracy for \texttt{Gemma-3-12b-it} trained with 20,000 training samples.}
    \label{fig:lora_rank}
    \vspace{-1.0em}
\end{figure}

\section{Experiments} \label{sec:experiments}
\subsection{Implementation Details} Below we provide details about configurations, hyperparameters, and training and evaluation setup.

\textit{Synthetic Data Generation:} For the input source and translation pairs used in synthetic data generation, we curate a dataset of $~3M$ data points from multiple public datasets, including \texttt{opus-100}, \texttt{opus-books}, \texttt{europarl}, \texttt{opus-tedtalks}, \texttt{tanzil}, \texttt{wikipedia}, \texttt{tatoeba}, and \texttt{WMT} 2019, 2017, and 2015. The dataset covers the following languages: Spanish (\texttt{es}), Russian (\texttt{ru}), German (\texttt{de}), Chinese (\texttt{zh}), English (\texttt{en}), Japanese (\texttt{ja}), and Hindi (\texttt{hi}). These languages span both high-resource and medium-resource settings, enabling diverse coverage for error span annotation generation. 

\begin{figure}
    \centering
    \includegraphics[width=\linewidth]{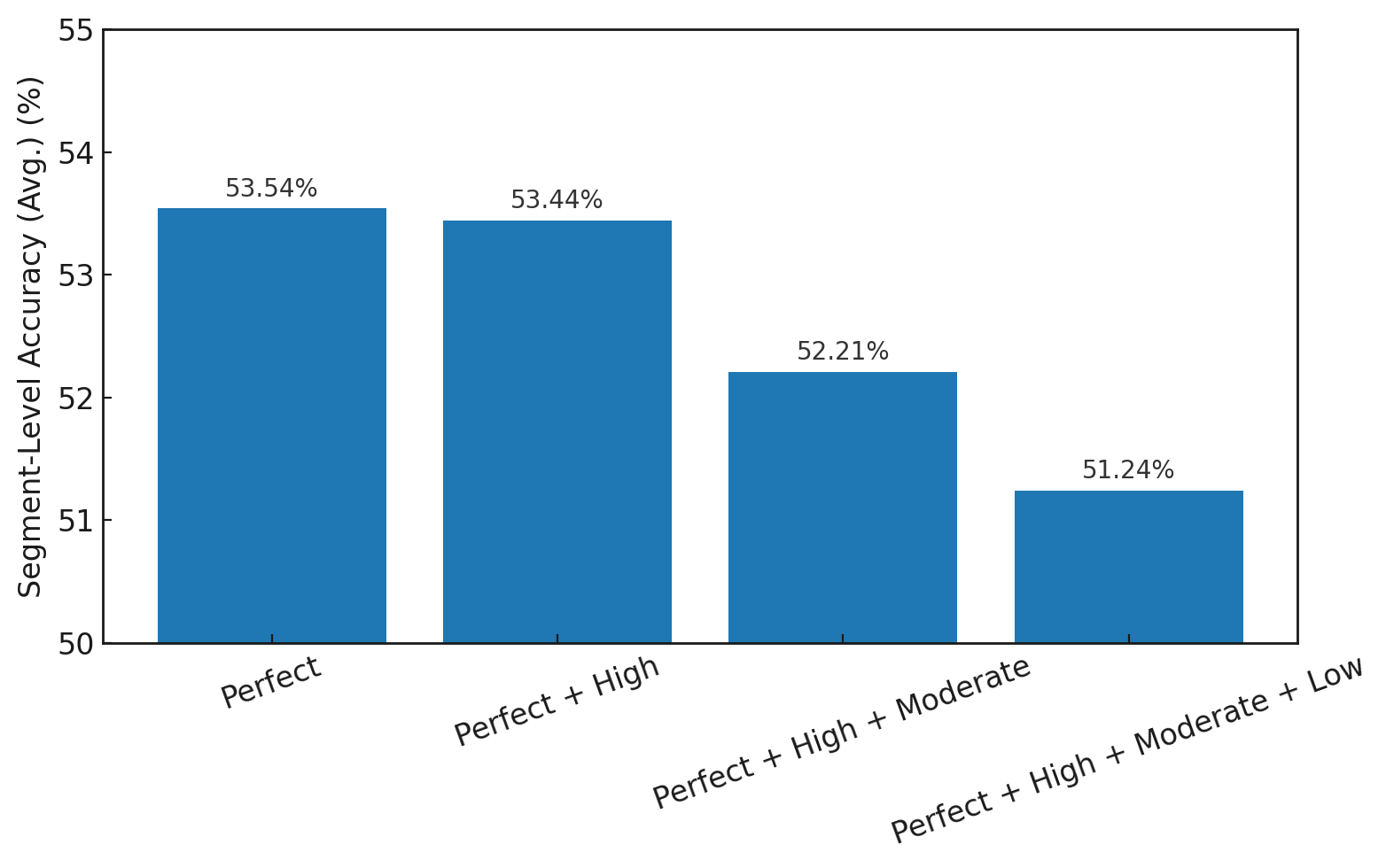}
    \caption{Impact of changing data quality through incorporating varying degrees of jury agreement in training data on the average segment level performance.}
    \label{fig:llm_agreement}
\end{figure}

For a set of LRMs $M$, we consider the following three models: \texttt{OpenAI o1} and \texttt{OpenAI o3}. We perform error-span aggregation using \texttt{GPT-4.1-mini} as a meta-judge. Unless stated otherwise, we report results for our fine-tuned models trained with MQM filtering based on \textit{Perfect-Agreement}, which comprises 99,214 (\textasciitilde 100K) high-quality synthetic samples. We finetune two small language models (\textit{Gemma-12B-it} and \textit{Gemma-3-4B-it}). Unless specified otherwise we perform LoRA based parameter efficient supervised finetuning with LoRA rank of 256.

\subsection{Discussion}

\vspace{0.1cm}
\noindent\textbf{Comparison against larger models.} In Table \ref{tab:tqlite_results}, we present results for distilled variants of TQLite. For \texttt{gemma-12b-it}, we observe that distillation consistently improves both segment-level and system-level performance over its pretrained counterpart, achieving an average segment-level accuracy of 55.03\%. We outperform all open LRM baselines from Table 2 which includes \texttt{qwen-3-8b-thinking}, \texttt{qwen-3-32b-thinking}, and \texttt{deepseek-r1-distill-llama-8b/70b}—by a substantial margin. While our results remain below the best closed LRM, we achieve them with significantly lower inference cost and latency, highlighting the benefits of Multi-LLM Jury based distillation for efficient translation quality judges.

\vspace{0.1cm}
\noindent\textbf{Impact of varying Data Quality.}
Jury agreement, as discussed earlier, is a good proxy for distillation data quality and provides an effective criterion for filtering samples during student finetuning. Figure \ref{fig:llm_agreement} illustrates this effect for the distilled \texttt{Gemma-3-12B-it} model trained on 20K samples. Training with only Perfect-Agreement data yields the highest accuracy (53.54\%), closely followed by Perfect + High agreement (53.44\%). Performance declines as lower-agreement samples are introduced, dropping to 52.21\% with Moderate and 51.24\% when Low agreement data is included. These results underscore that maintaining high jury agreement in training data is critical to preserving label quality and maximizing distillation effectiveness.

\begin{figure}
    \centering
    \includegraphics[width=\linewidth]{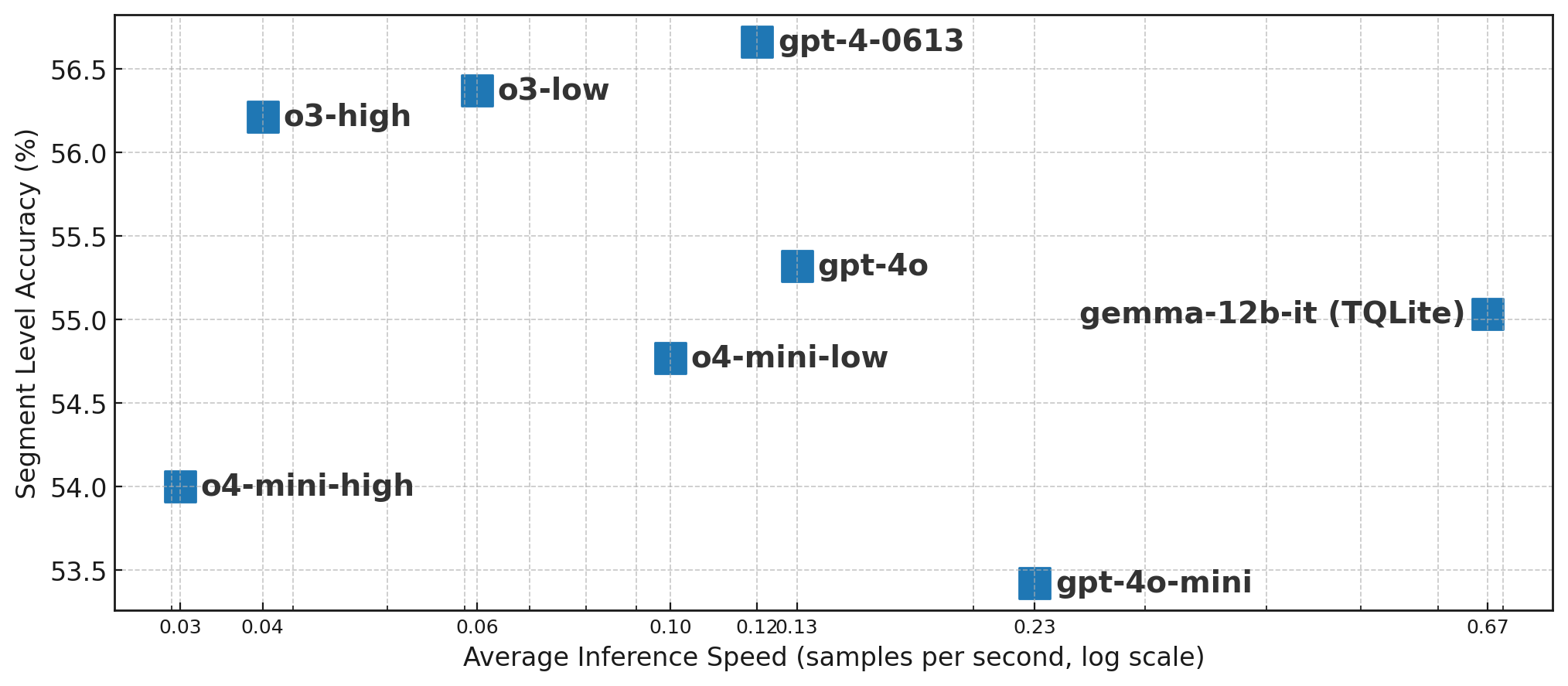}
    \caption{Inference time to performance tradeoff.}
    \label{fig:inferencetime}
    \vspace{-1.5em}
\end{figure}

\vspace{0.1cm}
\noindent \textbf{Impact of training samples.}
In Figure \ref{fig:distillation_scaling}, we present the effect of training set size on segment-level accuracy for the distilled \texttt{Gemma-3-12B-it} model. Accuracy improves consistently across the entire range, rising from 0.5259 at 5K samples to 0.5529 at ~99K. The largest single jump occurs between 10K (0.5159) and 20K (0.5354), indicating substantial benefit from moving out of the low-data regime. While gains per increment become smaller at larger scales, performance continues to increase steadily up to the maximum dataset size, showing limited signs of plateau within the tested range.

\vspace{0.1cm}
\noindent\textbf{Number of training parameters.}
In Figure \ref{fig:lora_rank}, we present the effect of LoRA rank—the number of trainable parameters—on segment-level accuracy for \texttt{Gemma-3-12B-it} trained with 20K samples. Accuracy rises steadily from rank 16 (52.91\%) to a peak at rank 128 (53.80\%), indicating that moderate capacity yields the best adaptation in this setting. The slight drop at rank 256 may stem from the 20K-sample training set being relatively small for such a high number of trainable parameters, limiting the model’s ability to fully exploit the added capacity and potentially introducing mild overfitting. Across all ranks, LoRA finetuning provides consistent improvements over the pretrained baseline.

\vspace{0.1cm}
\noindent\textbf{Inference time and model size.}
In Figure \ref{fig:inferencetime} we present the average segment level accuracy to inference speed trade-off for the best closed source models and TQLite (\texttt{gemma-12b-it}). As it can be observed that, TQLite has the best \textit{accuracy x inference time} metric, providing a good translation quality judgment at a significantly faster inference speed (lower inference time). This analysis was performed by using the APIs provided by the OpenAI for their various models, and using vLLM as the inference engine for \texttt{Gemma-3-12b-it}. We infer \texttt{Gemma-3-12b-it} using 4x A100 (40GB) GPUs.

\section{Conclusion}
In this work, we presented \texttt{TQLite}, a multi-LRM jury–guided distillation framework for efficient MQM-based translation quality evaluation. Our benchmarking showed that LLM evaluators are highly sensitive to prompt design, and that large reasoning models—especially in multi-LRM juries—deliver the strongest results but at high computational and monetary cost. \texttt{TQLite} distills agreement-filtered outputs from these juries into smaller open-source models, preserving much of their accuracy while enabling faster, cheaper inference. The resulting SLMs outperform all open baselines and approach closed frontier models, providing a scalable path to high-quality, cost-effective TQ evaluation.

\section*{Limitations}
Our study demonstrates the effectiveness of distilling multi-LRM juries to improve SLM performance for TQ evaluation; however, several limitations remain. First, SLM-based evaluators still fall short of the best LRM-based judges. Second, our synthetic training data—generated by an LRM jury, even with correct aggregation—may cause student models to inherit systematic biases from their teachers. Third, distilled SLM quality depends on LRM jury diversity and agreement. While high-agreement samples boost performance, filtering may reduce coverage of edge cases and harder translation errors. Lastly, our experiments target only high- and medium-resource language pairs from the WMT22 metrics test set; effectiveness for truly low-resource or morphologically complex languages remains untested. Further work is needed to assess cross-lingual transfer and generalization.

\section*{Ethics Statement}
We use only publicly available datasets that are free of personally identifiable or sensitive information.
Our approach relies on LLMs, which have a tendency to propagate biases present in their training data. 
We attempt to mitigate these systematic biases through aggregation, but some bias may remain.
Additionally, our approach focuses on training efficient SLMs, but LLMs are still notoriously resource-intensive for both training and inference.
Our distilled models are primarily evaluated on high- and medium-resource languages, so caution is advised when applying them to other settings or critical applications.
We aim to release code and data for transparency and reproducibility.

% \section*{Acknowledgements}
% TBD

\bibliography{anthology,custom}
\bibliographystyle{acl_natbib}

\appendix

\section{Experiment Details}
\label{sec:apx:exp_details}

Here, we outline all of the detailed experimental settings used across all of our baseline and \texttt{TQLite} experiments.
We begin with outlining settings for all benchmark experiments in \ref{sec:bench_laaj}, then cover settings for experiments with \texttt{TQLite} in the next section.

\subsection{Benchmarking Experiments}
\textbf{Baselines.}
All baseline performance metrics were obtained using the \href{https://github.com/google-research/mt-metrics-eval}{mt-metrics-eval} package, which contains the official scripts and scores for the WMT22 metrics task.
The accuracy for these baseline techniques match those found in prior work \cite{kocmi2023gemba}.

\textbf{Models.} The exact models that we use in our baseline experiments are as follows:
\begin{itemize}
    \item gpt-4 (\texttt{gpt-4-0613})
    \item gpt-4o (\texttt{gpt-4o-2024-08-06})
    \item gpt-4o-mini (\texttt{gpt-4o-mini-2024-07-18})
    \item o1 (\texttt{o1-2024-12-17})
    \item o3 (\texttt{o3-2025-04-16})
    \item o4-mini (\texttt{o4-mini-2025-04-16})
    \item gemini-2.5-pro
    \item gemini-2.5-flash
    \item \href{https://huggingface.co/google/gemma-3-4b-it}{gemma-3-4b-it}
    \item \href{https://huggingface.co/google/gemma-3-27b-it}{gemma-3-27b-it}
    \item \href{https://huggingface.co/Qwen/Qwen3-8B}{qwen-3-8b}
    \item \href{https://huggingface.co/Qwen/Qwen3-32B}{qwen-3-32b}
    \item \href{https://huggingface.co/deepseek-ai/DeepSeek-R1-Distill-Llama-8B}{deepseek-r1-distill-llama-8b}
    \item \href{https://huggingface.co/deepseek-ai/DeepSeek-R1-Distill-Llama-70B}{deepseek-r1-distill-llama-70b}
    \item \href{https://huggingface.co/meta-llama/Meta-Llama-3-8B-Instruct}{llama-3-8b-instruct}
    \item \href{https://huggingface.co/meta-llama/Meta-Llama-3-70B-Instruct}{llama-3-70b-instruct}
    \item \href{https://huggingface.co/CohereLabs/aya-expanse-8b}{aya-expanse-8b}
    \item \href{https://huggingface.co/CohereLabs/aya-expanse-32b}{aya-expanse-32b}
\end{itemize}

\textbf{Inference.} All OpenAI models are accessed directly via the OpenAI API, specifically via the \href{https://github.com/openai/openai-python}{associated Python library}.
Similarly, all (closed) Google models are accessed via the Vertex AI API by using the \texttt{google-cloud-aiplatform} SDK.
For all open models, we simply download the models via HuggingFace at the links provided above.

To run inference with open models, we use \href{https://pypi.org/project/vllm/}{vLLM} offline inference as demonstrated \href{https://docs.vllm.ai/en/stable/serving/offline_inference.html}{here}.
For the open reasoning models (i.e., qwen-3 and deepseek-r1-distill), we use \href{https://github.com/sgl-project/sglang}{SGLang} offline inference as demonstrated \href{https://docs.sglang.ai/basic_usage/offline_engine_api.html}{here}, which we find to be faster when running inference with long completions.
All $\sim$7B parameter models use a single 80Gb A100 GPU for inference, while all larger models use a $8\times$ 80Gb A100 node with a tensor parallel size of 8.

\textbf{Sampling.} Our default sampling settings across all models use a temperature of 0.0, a context length of 8,192, and max completion tokens of 4,096.
These sampling parameters match the settings of \cite{kocmi2023gemba}, and we use them whenever possible.
For all thinking (or reasoning) models, we increase the context length to 32,768 and the max completion tokens to 28,672\footnote{This is just the context length minus the maximum length of any prompt used in other experiments (4,096).}, which is the recommended setting from \href{https://platform.openai.com/docs/guides/reasoning}{OpenAI}.
We adopt the same setting for open reasoning models, though we find that the completions generated by these models are typically much shorter (i.e., $\sim$ 1-2K tokens or less). 

qwen-3 models are extremely sensitive to sampling hyperparameters, and the setting must be changed depending on whether thinking is enabled or not.
During initial experiments outlined in Figure \ref{fig:gemba_mqm}, we set thinking to be disabled in both qwen-3 models and use the settings recommended \href{https://huggingface.co/Qwen/Qwen3-32B}{here}:
\begin{itemize}
    \item Temperature = 0.7
    \item TopP = 0.8
    \item TopK = 20
    \item MinP = 0
\end{itemize}
Similarly, when these models are used with thinking enabled in Table \ref{tab:llm_baselines}, we update the temperature to a value of 0.6 and TopP to a value of 0.95 (i.e., the recommended sampling parameters for thinking).

For the deepseek-r1-distill reasoning models, we use the context window for reasoning models outlined above and the following sampling settings:
\begin{itemize}
    \item Temperature = 0.6
    \item TopP = 0.95
    \item TopK = 20
    \item MinP = 0.1
\end{itemize}
These are the recommended settings from the \href{https://huggingface.co/deepseek-ai/DeepSeek-R1-Distill-Qwen-32B}{model provider}.
OpenAI reasoning models like o1, o3, and o4-mini do not allow for the sampling temperature to be set by the user---\emph{the temperature is fixed to a value of one}.

\section{Prompt Templates}
\label{sec:apx:prompts}

\begin{figure}
    \centering
    \includegraphics[width=1.0\linewidth]{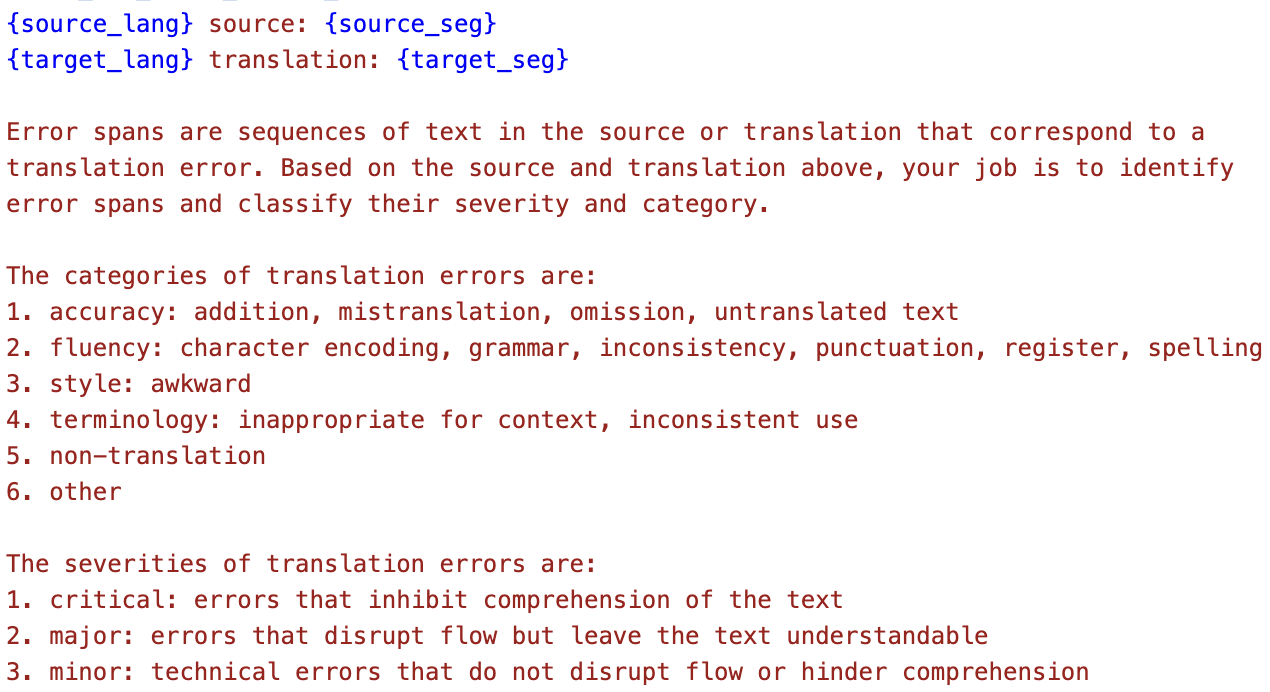}
    \caption{Our \texttt{TQLite} prompt, which is a slightly modified version of the prompt used by \texttt{Gemba-MQM} \cite{kocmi2023gemba}.}
    \label{fig:tqlite_prompt}
\end{figure}

We now outline each of the prompt templates used in both benchmarking experiments in Section \ref{sec:bench_laaj} and \texttt{TQLite} experiments in Section \ref{sec:experiments}.
The original \texttt{Gemba-MQM} prompt is taken directly from \cite{kocmi2023gemba}.
All models use their default chat template.
We make no changes to the tokenizer or chat template for any model. 
For all models that accept a system prompt, we also include the original system prompt from \texttt{Gemba-MQM}.

\textbf{Few-shot examples.}
For all experiments in Section \ref{sec:bench_laaj}, we use the same three few-shot examples used by \texttt{Gemba-MQM} \citeauthor{kocmi2023gemba}.
We directly take these examples from the \href{https://github.com/MicrosoftTranslator/GEMBA}{code repository} for \texttt{Gemba-MQM}.
For standard prompts, these few-shot examples are passed as multiple turns of a chat, where each user chat turn receives the entire prompt with the actual few-shot example inserted.
The associated, ground truth error spans for each few-shot example are then returned as an assistant turn in the chat.
The formatting for few-shot examples in single-turn prompts is outlined below.

Our modified \texttt{TQLite} prompt is shown in Figure \ref{fig:tqlite_prompt}.
Based on this \texttt{TQLite} prompt, we create two variants of the prompt that induce either structure text or JSON output formatting by adding the following instructions to the prompt:
\begin{tcolorbox}[colback=gray!10, colframe=black, title=Formatting Instruction]
\textbf{Structured Text:}\texttt{ Please provide your error spans below. Your error spans should be organized under three headers: "Critical:", "Major:" and "Minor:". Under each header, provide your error span(s) for that severity. Each error span must include both the category of error and the error span itself separated by a hyphen with the format "<category> - <span>". Please output your error spans only with no additional explanation.}\\\\
\textbf{JSON:}\texttt{ Please output all of your error spans as a json document. Please output only the json document alone with nothing else. Be sure to surround any quote from the source or translation with single quotes; e.g., 'xxx'.}
\end{tcolorbox}
For prompts with structured outputs, we use the same \texttt{TQLite} prompt outlined above, and these additional formatting instructions are simply appended to the end of the prompt.

\begin{figure}
    \centering
    \includegraphics[width=1.0\linewidth]{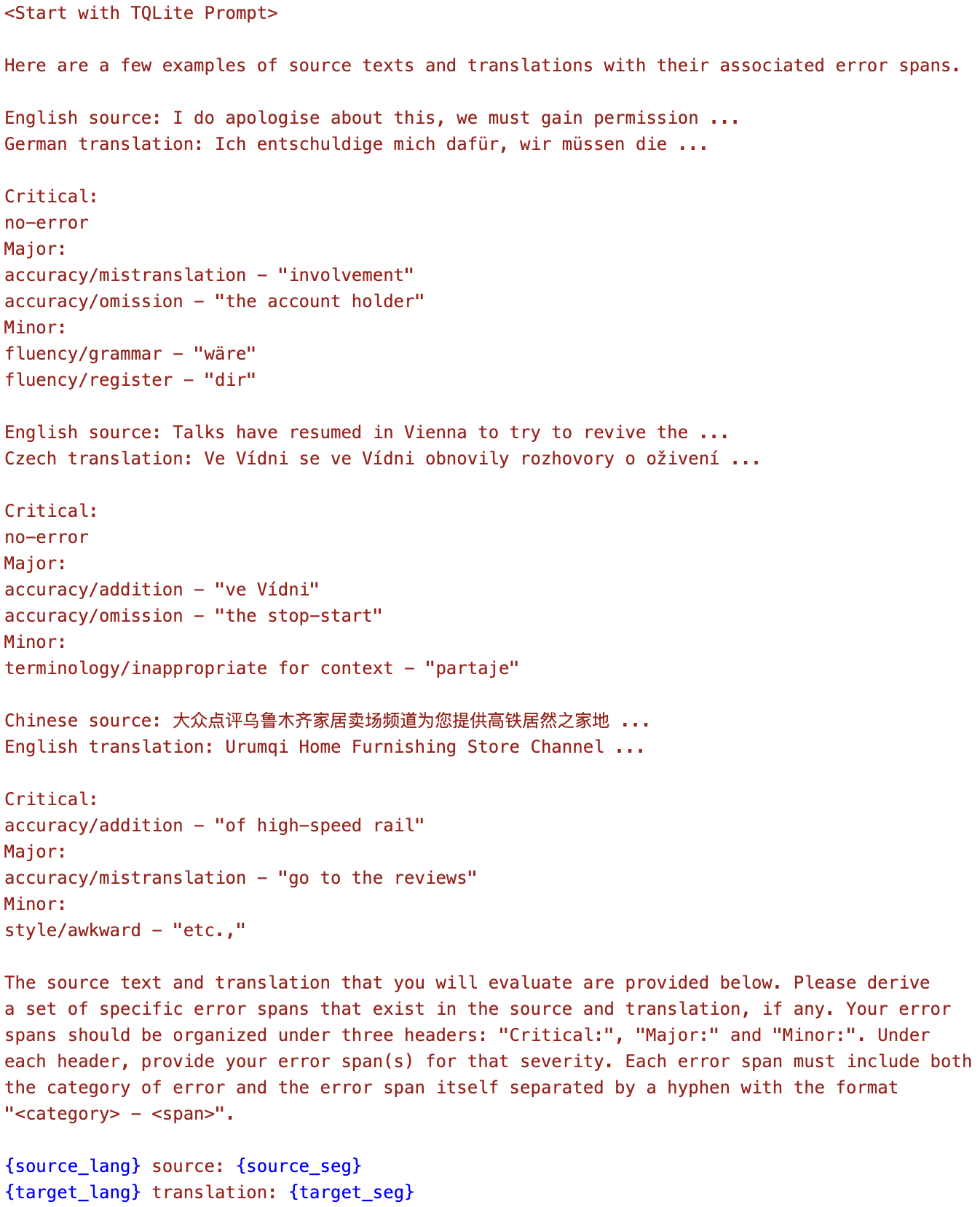}
    \caption{Our \texttt{TQLite-ST} prompt, which modifies the \texttt{TQLite} prompt to fully reside in a single chat turn.}
    \label{fig:tqlite_st_prompt}
\end{figure}

Our single turn prompt \texttt{TQLite-ST} is provided in Figure \ref{fig:tqlite_st_prompt}.
This same prompt is used by both standard LLMs and reasoning models. 
We also create a variant of this prompt that modifies the end of the single-turn prompt to include a zero-shot CoT prompt \cite{kojima2022large} to elicit an explanation from standard LLMs along with their evaluation:
\begin{tcolorbox}[colback=gray!10, colframe=black, title=Zero-Shot CoT Instruction]
\texttt{ Let's think step-by-step. The source text and translation that you will evaluate are provided below. Begin by outputting an explanation of the errors that may or may not exist in the source and translation. Once you have fully explained the translation errors, please derive a set of specific error spans that exist in the source and translation, if any. <Insert Structured Text Instruction>}\\\\
\texttt{\{source\_lang\} source: \{source\_seg\}}\\
\texttt{\{tgt\_lang\} translation: \{tgt\_seg\}}
\end{tcolorbox}

\textbf{Output parsing.}
To parse the output of our models and generate the final MQM scores, we use the code base from MQM-APE \cite{lu2024mqm} found \href{https://github.com/Coldmist-Lu/MQM_APE}{here}.
The post-processing code in the MQM portion of this repository (i.e., excluding any of the post-editing code) handles parsing the structured textual output format for TQ evaluation.
However, we slightly modify this code to make it more lenient when processing errors spans.
Specifically, in the \texttt{<category> - <span>} structure, we allow the span portion of the error to include any text, rather than forcing the model to output a direct quote and ignoring the line otherwise.

For parsing JSON outputs, we simply strip any surrounding JSON formatting from the model's output and load the JSON document using the JSON package in python.
If an error occurs when loading the JSON, we simply set the score for that segment as \texttt{None}.
For LLMs and reasoning models, JSON parsing errors occur in a very small number of cases (i.e., $\ll 1\%$ of segments).
For SLMs, this ratio can be slightly higher, though it is still small (i.e., $1-2\%$ of segments).

\begin{tcolorbox}[
  title=Data Generation Prompt,
  colback=gray!10, colframe=black,
  width=\textwidth,     % full text width
  breakable,            % allow page breaks
  enhanced,
  listing engine=listings,
  listing only,
  listing options={
    basicstyle=\ttfamily\small,
    breaklines=true, breakatwhitespace=false,
    columns=fullflexible,              % better wrapping of long tokens
    keepspaces=true,                   % preserve spaces/indentation
    showstringspaces=false
  }
]
Basedwe on the source segment and machine translation below, identify error spans in the translation and classify them for their severity and category. You would also explain why you think that its an error span.

The categories of errors are: 
1. accuracy (addition, mistranslation, omission, untranslated text), 
2. fluency (character encoding, grammar, inconsistency, punctuation, register, spelling), 
3. style (awkward), 
4. terminology (inappropriate for context, inconsistent use), 
5. non-translation, 
6. other
7. no-error.

Each error is classified as one of three severities: critical, major, and minor. 
1. Critical Error: Critical errors inhibit comprehension of the text. 
2. Major Error: Major errors disrupt the flow, but what the text is trying to say is still understandable. 
3. Minor Error: Minor errors are technically errors, but do not disrupt the flow or hinder comprehension."""

Here are few examples:
<Few Set of Examples>

Below is src\_lang source text and tgt\_lang machine translation pair that you are suppose to evaluate. Give the full answer json as output.

Input:
\{\\
"source\_lang": {src\_lang}, \\
"source\_sentence": {src\_sentence}, \\ 
"target\_lang": {tgt\_lang}, \\
"target\_sentence": {tgt\_sentence} \\
\}\\

Output:
\end{tcolorbox}

\end{document}

%% file: tables/tqlite_distilled_table.tex
\begin{table*}[ht]
\footnotesize
\centering
\begin{tabular}{clcccccc}
\toprule
\multirow{2}{*}{\textbf{Category}} & \multirow{2}{*}{\textbf{Model}} &
\multicolumn{2}{c}{\textbf{System-Level Acc.}} &
\multicolumn{4}{c}{\textbf{Segment-Level Acc*}} \\
\cmidrule(lr){3-4} \cmidrule(lr){5-8}
& &
\multicolumn{2}{c}{\textbf{All LPs}} &
\multicolumn{1}{c}{\textbf{En-De}} &
\multicolumn{1}{c}{\textbf{En-Ru}} &
\multicolumn{1}{c}{\textbf{Zh-En}} &
\multicolumn{1}{c}{\textbf{Avg.}} \\
\toprule
\multirow{2}{*}{\shortstack{Pretrained}} & \texttt{gemma-3-4b-it} & \multicolumn{2}{c}{86.50\%} & 55.91\% & 52.78\% & 51.41\% & 53.36\% \\
 &\texttt{gemma-3-12b-it} & \multicolumn{2}{c}{78.10\%} & 56.04\% & 52.38\% &	49.48\%	& 52.63\% \\
\midrule
\multirow{2}{*}{\shortstack{TQLite-Distilled}} & \texttt{gemma-3-4b-it} & \multicolumn{2}{c}{82.48\%} & 56.20\% & 53.81\% & 49.40\% & 53.13\% \\
 &\texttt{gemma-3-12b-it} & \multicolumn{2}{c}{85.04\%} & 57.57\% & 55.30\% &	52.22\%	& 55.03\% \\
\bottomrule
\end{tabular}
\caption{Performance of LRMs on the WMT22 test set using the \texttt{TQLite-ST} prompt with structured text outputs. }
\label{tab:tqlite_results}
\end{table*}

%% file: emnlp2023.bbl
\begin{thebibliography}{59}
\expandafter\ifx\csname natexlab\endcsname\relax\def\natexlab#1{#1}\fi

\bibitem[{Achiam et~al.(2023)Achiam, Adler, Agarwal, Ahmad, Akkaya, Aleman, Almeida, Altenschmidt, Altman, Anadkat et~al.}]{achiam2023gpt}
Josh Achiam, Steven Adler, Sandhini Agarwal, Lama Ahmad, Ilge Akkaya, Florencia~Leoni Aleman, Diogo Almeida, Janko Altenschmidt, Sam Altman, Shyamal Anadkat, et~al. 2023.
\newblock Gpt-4 technical report.
\newblock \emph{arXiv preprint arXiv:2303.08774}.

\bibitem[{Anil et~al.(2023)Anil, Dai, Firat, Johnson, Lepikhin, Passos, Shakeri, Taropa, Bailey, Chen et~al.}]{anil2023palm}
Rohan Anil, Andrew~M Dai, Orhan Firat, Melvin Johnson, Dmitry Lepikhin, Alexandre Passos, Siamak Shakeri, Emanuel Taropa, Paige Bailey, Zhifeng Chen, et~al. 2023.
\newblock Palm 2 technical report.
\newblock \emph{arXiv preprint arXiv:2305.10403}.

\bibitem[{Banerjee and Lavie(2005)}]{banerjee2005meteor}
Satanjeev Banerjee and Alon Lavie. 2005.
\newblock Meteor: An automatic metric for mt evaluation with improved correlation with human judgments.
\newblock In \emph{Proceedings of the acl workshop on intrinsic and extrinsic evaluation measures for machine translation and/or summarization}, pages 65--72.

\bibitem[{Brown et~al.(2020)Brown, Mann, Ryder, Subbiah, Kaplan, Dhariwal, Neelakantan, Shyam, Sastry, Askell et~al.}]{brown2020language}
Tom Brown, Benjamin Mann, Nick Ryder, Melanie Subbiah, Jared~D Kaplan, Prafulla Dhariwal, Arvind Neelakantan, Pranav Shyam, Girish Sastry, Amanda Askell, et~al. 2020.
\newblock Language models are few-shot learners.
\newblock \emph{Advances in neural information processing systems}, 33:1877--1901.

\bibitem[{Chiang and Lee(2023)}]{chiang2023can}
Cheng-Han Chiang and Hung-yi Lee. 2023.
\newblock Can large language models be an alternative to human evaluations?
\newblock \emph{arXiv preprint arXiv:2305.01937}.

\bibitem[{Comanici et~al.(2025)Comanici, Bieber, Schaekermann, Pasupat, Sachdeva, Dhillon, Blistein, Ram, Zhang, Rosen et~al.}]{comanici2025gemini}
Gheorghe Comanici, Eric Bieber, Mike Schaekermann, Ice Pasupat, Noveen Sachdeva, Inderjit Dhillon, Marcel Blistein, Ori Ram, Dan Zhang, Evan Rosen, et~al. 2025.
\newblock Gemini 2.5: Pushing the frontier with advanced reasoning, multimodality, long context, and next generation agentic capabilities.
\newblock \emph{arXiv preprint arXiv:2507.06261}.

\bibitem[{Dang et~al.(2024)Dang, Singh, D'souza, Ahmadian, Salamanca, Smith, Peppin, Hong, Govindassamy, Zhao et~al.}]{dang2024aya}
John Dang, Shivalika Singh, Daniel D'souza, Arash Ahmadian, Alejandro Salamanca, Madeline Smith, Aidan Peppin, Sungjin Hong, Manoj Govindassamy, Terrence Zhao, et~al. 2024.
\newblock Aya expanse: Combining research breakthroughs for a new multilingual frontier.
\newblock \emph{arXiv preprint arXiv:2412.04261}.

\bibitem[{Deutsch et~al.(2023)Deutsch, Foster, and Freitag}]{deutsch2023ties}
Daniel Deutsch, George~F Foster, and Markus Freitag. 2023.
\newblock Ties matter: Modifying kendall's tau for modern metric meta-evaluation.
\newblock \emph{CoRR}.

\bibitem[{Dubey et~al.(2024)Dubey, Jauhri, Pandey, Kadian, Al-Dahle, Letman, Mathur, Schelten, Yang, Fan et~al.}]{dubey2024llama}
Abhimanyu Dubey, Abhinav Jauhri, Abhinav Pandey, Abhishek Kadian, Ahmad Al-Dahle, Aiesha Letman, Akhil Mathur, Alan Schelten, Amy Yang, Angela Fan, et~al. 2024.
\newblock The llama 3 herd of models.
\newblock \emph{arXiv e-prints}, pages arXiv--2407.

\bibitem[{Feng et~al.(2025)Feng, Fang, Ma, and Wang}]{feng2025efficient}
Sicheng Feng, Gongfan Fang, Xinyin Ma, and Xinchao Wang. 2025.
\newblock Efficient reasoning models: A survey.
\newblock \emph{arXiv preprint arXiv:2504.10903}.

\bibitem[{Fernandes et~al.(2023)Fernandes, Deutsch, Finkelstein, Riley, Martins, Neubig, Garg, Clark, Freitag, and Firat}]{fernandes2023devil}
Patrick Fernandes, Daniel Deutsch, Mara Finkelstein, Parker Riley, Andr{\'e}~FT Martins, Graham Neubig, Ankush Garg, Jonathan~H Clark, Markus Freitag, and Orhan Firat. 2023.
\newblock The devil is in the errors: Leveraging large language models for fine-grained machine translation evaluation.
\newblock \emph{arXiv preprint arXiv:2308.07286}.

\bibitem[{Freitag et~al.(2021)Freitag, Foster, Grangier, Ratnakar, Tan, and Macherey}]{freitag2021experts}
Markus Freitag, George Foster, David Grangier, Viresh Ratnakar, Qijun Tan, and Wolfgang Macherey. 2021.
\newblock Experts, errors, and context: A large-scale study of human evaluation for machine translation.
\newblock \emph{Transactions of the Association for Computational Linguistics}, 9:1460--1474.

\bibitem[{Freitag et~al.(2024)Freitag, Mathur, Deutsch, Lo, Avramidis, Rei, Thompson, Blain, Kocmi, Wang et~al.}]{freitag2024llms}
Markus Freitag, Nitika Mathur, Daniel Deutsch, Chi-Kiu Lo, Eleftherios Avramidis, Ricardo Rei, Brian Thompson, Frederic Blain, Tom Kocmi, Jiayi Wang, et~al. 2024.
\newblock Are llms breaking mt metrics? results of the wmt24 metrics shared task.
\newblock In \emph{Proceedings of the Ninth Conference on Machine Translation}, pages 47--81.

\bibitem[{Freitag et~al.(2022)Freitag, Rei, Mathur, Lo, Stewart, Avramidis, Kocmi, Foster, Lavie, and Martins}]{freitag2022results}
Markus Freitag, Ricardo Rei, Nitika Mathur, Chi-kiu Lo, Craig Stewart, Eleftherios Avramidis, Tom Kocmi, George Foster, Alon Lavie, and Andr{\'e}~FT Martins. 2022.
\newblock Results of wmt22 metrics shared task: Stop using bleu--neural metrics are better and more robust.
\newblock In \emph{Proceedings of the Seventh Conference on Machine Translation (WMT)}, pages 46--68.

\bibitem[{Fu et~al.(2023)Fu, Ng, Jiang, and Liu}]{fu2023gptscore}
Jinlan Fu, See-Kiong Ng, Zhengbao Jiang, and Pengfei Liu. 2023.
\newblock Gptscore: Evaluate as you desire.
\newblock \emph{arXiv preprint arXiv:2302.04166}.

\bibitem[{Gu et~al.(2024)Gu, Jiang, Shi, Tan, Zhai, Xu, Li, Shen, Ma, Liu et~al.}]{gu2024survey}
Jiawei Gu, Xuhui Jiang, Zhichao Shi, Hexiang Tan, Xuehao Zhai, Chengjin Xu, Wei Li, Yinghan Shen, Shengjie Ma, Honghao Liu, et~al. 2024.
\newblock A survey on llm-as-a-judge.
\newblock \emph{arXiv preprint arXiv:2411.15594}.

\bibitem[{Guerreiro et~al.(2024)Guerreiro, Rei, Stigt, Coheur, Colombo, and Martins}]{guerreiro2024xcomet}
Nuno~M Guerreiro, Ricardo Rei, Daan~van Stigt, Luisa Coheur, Pierre Colombo, and Andr{\'e}~FT Martins. 2024.
\newblock xcomet: Transparent machine translation evaluation through fine-grained error detection.
\newblock \emph{Transactions of the Association for Computational Linguistics}, 12:979--995.

\bibitem[{Guo et~al.(2025)Guo, Yang, Zhang, Song, Zhang, Xu, Zhu, Ma, Wang, Bi et~al.}]{guo2025deepseek}
Daya Guo, Dejian Yang, Haowei Zhang, Junxiao Song, Ruoyu Zhang, Runxin Xu, Qihao Zhu, Shirong Ma, Peiyi Wang, Xiao Bi, et~al. 2025.
\newblock Deepseek-r1: Incentivizing reasoning capability in llms via reinforcement learning.
\newblock \emph{arXiv preprint arXiv:2501.12948}.

\bibitem[{Hinton et~al.(2015)Hinton, Vinyals, and Dean}]{hinton2015distilling}
Geoffrey Hinton, Oriol Vinyals, and Jeff Dean. 2015.
\newblock Distilling the knowledge in a neural network.
\newblock \emph{arXiv preprint arXiv:1503.02531}.

\bibitem[{Hoffmann et~al.(2022)Hoffmann, Borgeaud, Mensch, Buchatskaya, Cai, Rutherford, Casas, Hendricks, Welbl, Clark et~al.}]{hoffmann2022training}
Jordan Hoffmann, Sebastian Borgeaud, Arthur Mensch, Elena Buchatskaya, Trevor Cai, Eliza Rutherford, Diego de~Las Casas, Lisa~Anne Hendricks, Johannes Welbl, Aidan Clark, et~al. 2022.
\newblock Training compute-optimal large language models.
\newblock \emph{arXiv preprint arXiv:2203.15556}.

\bibitem[{Hurst et~al.(2024)Hurst, Lerer, Goucher, Perelman, Ramesh, Clark, Ostrow, Welihinda, Hayes, Radford et~al.}]{hurst2024gpt}
Aaron Hurst, Adam Lerer, Adam~P Goucher, Adam Perelman, Aditya Ramesh, Aidan Clark, AJ~Ostrow, Akila Welihinda, Alan Hayes, Alec Radford, et~al. 2024.
\newblock Gpt-4o system card.
\newblock \emph{arXiv preprint arXiv:2410.21276}.

\bibitem[{Jaech et~al.(2024)Jaech, Kalai, Lerer, Richardson, El-Kishky, Low, Helyar, Madry, Beutel, Carney et~al.}]{jaech2024openai}
Aaron Jaech, Adam Kalai, Adam Lerer, Adam Richardson, Ahmed El-Kishky, Aiden Low, Alec Helyar, Aleksander Madry, Alex Beutel, Alex Carney, et~al. 2024.
\newblock Openai o1 system card.
\newblock \emph{arXiv preprint arXiv:2412.16720}.

\bibitem[{Juraska et~al.(2024)Juraska, Deutsch, Finkelstein, and Freitag}]{juraska2024metricx}
Juraj Juraska, Daniel Deutsch, Mara Finkelstein, and Markus Freitag. 2024.
\newblock Metricx-24: The google submission to the wmt 2024 metrics shared task.
\newblock \emph{arXiv preprint arXiv:2410.03983}.

\bibitem[{Kaplan et~al.(2020)Kaplan, McCandlish, Henighan, Brown, Chess, Child, Gray, Radford, Wu, and Amodei}]{kaplan2020scaling}
Jared Kaplan, Sam McCandlish, Tom Henighan, Tom~B Brown, Benjamin Chess, Rewon Child, Scott Gray, Alec Radford, Jeffrey Wu, and Dario Amodei. 2020.
\newblock Scaling laws for neural language models.
\newblock \emph{arXiv preprint arXiv:2001.08361}.

\bibitem[{Kim et~al.(2023)Kim, Shin, Cho, Jang, Longpre, Lee, Yun, Shin, Kim, Thorne et~al.}]{kim2023prometheus}
Seungone Kim, Jamin Shin, Yejin Cho, Joel Jang, Shayne Longpre, Hwaran Lee, Sangdoo Yun, Seongjin Shin, Sungdong Kim, James Thorne, et~al. 2023.
\newblock Prometheus: Inducing fine-grained evaluation capability in language models.
\newblock In \emph{The Twelfth International Conference on Learning Representations}.

\bibitem[{Kim et~al.(2024)Kim, Suk, Longpre, Lin, Shin, Welleck, Neubig, Lee, Lee, and Seo}]{kim2024prometheus}
Seungone Kim, Juyoung Suk, Shayne Longpre, Bill~Yuchen Lin, Jamin Shin, Sean Welleck, Graham Neubig, Moontae Lee, Kyungjae Lee, and Minjoon Seo. 2024.
\newblock Prometheus 2: An open source language model specialized in evaluating other language models.
\newblock \emph{arXiv preprint arXiv:2405.01535}.

\bibitem[{Kocmi et~al.(2022)Kocmi, Bawden, Bojar, Dvorkovich, Federmann, Fishel, Gowda, Graham, Grundkiewicz, Haddow et~al.}]{kocmi2022findings}
Tom Kocmi, Rachel Bawden, Ond{\v{r}}ej Bojar, Anton Dvorkovich, Christian Federmann, Mark Fishel, Thamme Gowda, Yvette Graham, Roman Grundkiewicz, Barry Haddow, et~al. 2022.
\newblock Findings of the 2022 conference on machine translation (wmt22).
\newblock In \emph{Proceedings of the Seventh Conference on Machine Translation (WMT)}, pages 1--45.

\bibitem[{Kocmi and Federmann(2023{\natexlab{a}})}]{kocmi2023gemba}
Tom Kocmi and Christian Federmann. 2023{\natexlab{a}}.
\newblock Gemba-mqm: Detecting translation quality error spans with gpt-4.
\newblock \emph{arXiv preprint arXiv:2310.13988}.

\bibitem[{Kocmi and Federmann(2023{\natexlab{b}})}]{kocmi2023large}
Tom Kocmi and Christian Federmann. 2023{\natexlab{b}}.
\newblock Large language models are state-of-the-art evaluators of translation quality.
\newblock \emph{arXiv preprint arXiv:2302.14520}.

\bibitem[{Kocmi et~al.(2021)Kocmi, Federmann, Grundkiewicz, Junczys-Dowmunt, Matsushita, and Menezes}]{kocmi2021ship}
Tom Kocmi, Christian Federmann, Roman Grundkiewicz, Marcin Junczys-Dowmunt, Hitokazu Matsushita, and Arul Menezes. 2021.
\newblock To ship or not to ship: An extensive evaluation of automatic metrics for machine translation.
\newblock \emph{arXiv preprint arXiv:2107.10821}.

\bibitem[{Kojima et~al.(2022)Kojima, Gu, Reid, Matsuo, and Iwasawa}]{kojima2022large}
Takeshi Kojima, Shixiang~Shane Gu, Machel Reid, Yutaka Matsuo, and Yusuke Iwasawa. 2022.
\newblock Large language models are zero-shot reasoners.
\newblock \emph{Advances in neural information processing systems}, 35:22199--22213.

\bibitem[{Lambert et~al.(2024)Lambert, Morrison, Pyatkin, Huang, Ivison, Brahman, Miranda, Liu, Dziri, Lyu et~al.}]{lambert2024tulu}
Nathan Lambert, Jacob Morrison, Valentina Pyatkin, Shengyi Huang, Hamish Ivison, Faeze Brahman, Lester James~V Miranda, Alisa Liu, Nouha Dziri, Shane Lyu, et~al. 2024.
\newblock Tulu 3: Pushing frontiers in open language model post-training.
\newblock \emph{arXiv preprint arXiv:2411.15124}.

\bibitem[{Li et~al.(2023)Li, Zhang, Dubois, Taori, Gulrajani, Guestrin, Liang, and Hashimoto}]{li2023alpacaeval}
Xuechen Li, Tianyi Zhang, Yann Dubois, Rohan Taori, Ishaan Gulrajani, Carlos Guestrin, Percy Liang, and Tatsunori~B Hashimoto. 2023.
\newblock Alpacaeval: An automatic evaluator of instruction-following models.

\bibitem[{Liu et~al.(2023)Liu, Iter, Xu, Wang, Xu, and Zhu}]{liu2023g}
Yang Liu, Dan Iter, Yichong Xu, Shuohang Wang, Ruochen Xu, and Chenguang Zhu. 2023.
\newblock G-eval: Nlg evaluation using gpt-4 with better human alignment.
\newblock \emph{arXiv preprint arXiv:2303.16634}.

\bibitem[{Lommel et~al.(2014)Lommel, Uszkoreit, and Burchardt}]{lommel2014multidimensional}
Arle Lommel, Hans Uszkoreit, and Aljoscha Burchardt. 2014.
\newblock Multidimensional quality metrics (mqm): A framework for declaring and describing translation quality metrics.
\newblock \emph{Tradum{\`a}tica}, (12):0455--463.

\bibitem[{Lu et~al.(2024)Lu, Ding, Zhang, Zhang, and Tao}]{lu2024mqm}
Qingyu Lu, Liang Ding, Kanjian Zhang, Jinxia Zhang, and Dacheng Tao. 2024.
\newblock Mqm-ape: Toward high-quality error annotation predictors with automatic post-editing in llm translation evaluators.
\newblock \emph{arXiv preprint arXiv:2409.14335}.

\bibitem[{Lu et~al.(2023)Lu, Qiu, Ding, Xie, and Tao}]{lu2023error}
Qingyu Lu, Baopu Qiu, Liang Ding, Liping Xie, and Dacheng Tao. 2023.
\newblock Error analysis prompting enables human-like translation evaluation in large language models: A case study on chatgpt.

\bibitem[{Mathur et~al.(2020)Mathur, Baldwin, and Cohn}]{mathur2020tangled}
Nitika Mathur, Timothy Baldwin, and Trevor Cohn. 2020.
\newblock Tangled up in bleu: Reevaluating the evaluation of automatic machine translation evaluation metrics.
\newblock \emph{arXiv preprint arXiv:2006.06264}.

\bibitem[{Meta(2025)}]{meta2025llama4}
Meta. 2025.
\newblock \href {https://ai.meta.com/blog/llama-4-multimodal-intelligence/} {The llama 4 herd: The beginning of a new era of natively multimodal ai innovation}.
\newblock Accessed: 2025-08-12.

\bibitem[{OpenAI(2025)}]{openai2025o3}
OpenAI. 2025.
\newblock \href {https://openai.com/index/introducing-o3-and-o4-mini/} {Introducing openai o3 and o4-mini}.
\newblock Accessed: 2025-08-12.

\bibitem[{Papineni et~al.(2002)Papineni, Roukos, Ward, and Zhu}]{papineni2002bleu}
Kishore Papineni, Salim Roukos, Todd Ward, and Wei-Jing Zhu. 2002.
\newblock Bleu: a method for automatic evaluation of machine translation.
\newblock In \emph{Proceedings of the 40th annual meeting of the Association for Computational Linguistics}, pages 311--318.

\bibitem[{Rei et~al.(2022{\natexlab{a}})Rei, De~Souza, Alves, Zerva, Farinha, Glushkova, Lavie, Coheur, and Martins}]{rei2022comet}
Ricardo Rei, Jos{\'e}~GC De~Souza, Duarte Alves, Chrysoula Zerva, Ana~C Farinha, Taisiya Glushkova, Alon Lavie, Luisa Coheur, and Andr{\'e}~FT Martins. 2022{\natexlab{a}}.
\newblock Comet-22: Unbabel-ist 2022 submission for the metrics shared task.
\newblock In \emph{Proceedings of the Seventh Conference on Machine Translation (WMT)}, pages 578--585.

\bibitem[{Rei et~al.(2021)Rei, Farinha, Zerva, Van~Stigt, Stewart, Ramos, Glushkova, Martins, and Lavie}]{rei2021references}
Ricardo Rei, Ana~C Farinha, Chrysoula Zerva, Daan Van~Stigt, Craig Stewart, Pedro Ramos, Taisiya Glushkova, Andr{\'e}~FT Martins, and Alon Lavie. 2021.
\newblock Are references really needed? unbabel-ist 2021 submission for the metrics shared task.
\newblock In \emph{Proceedings of the Sixth Conference on Machine Translation}, pages 1030--1040.

\bibitem[{Rei et~al.(2022{\natexlab{b}})Rei, Treviso, Guerreiro, Zerva, Farinha, Maroti, De~Souza, Glushkova, Alves, Lavie et~al.}]{rei2022cometkiwi}
Ricardo Rei, Marcos Treviso, Nuno~M Guerreiro, Chrysoula Zerva, Ana~C Farinha, Christine Maroti, Jos{\'e}~GC De~Souza, Taisiya Glushkova, Duarte~M Alves, Alon Lavie, et~al. 2022{\natexlab{b}}.
\newblock Cometkiwi: Ist-unbabel 2022 submission for the quality estimation shared task.
\newblock \emph{arXiv preprint arXiv:2209.06243}.

\bibitem[{Sellam et~al.(2020)Sellam, Das, and Parikh}]{sellam2020bleurt}
Thibault Sellam, Dipanjan Das, and Ankur~P Parikh. 2020.
\newblock Bleurt: Learning robust metrics for text generation.
\newblock \emph{arXiv preprint arXiv:2004.04696}.

\bibitem[{Shao et~al.(2024)Shao, Wang, Zhu, Xu, Song, Bi, Zhang, Zhang, Li, Wu et~al.}]{shao2024deepseekmath}
Zhihong Shao, Peiyi Wang, Qihao Zhu, Runxin Xu, Junxiao Song, Xiao Bi, Haowei Zhang, Mingchuan Zhang, YK~Li, Yang Wu, et~al. 2024.
\newblock Deepseekmath: Pushing the limits of mathematical reasoning in open language models.
\newblock \emph{arXiv preprint arXiv:2402.03300}.

\bibitem[{Stiennon et~al.(2020)Stiennon, Ouyang, Wu, Ziegler, Lowe, Voss, Radford, Amodei, and Christiano}]{stiennon2020learning}
Nisan Stiennon, Long Ouyang, Jeffrey Wu, Daniel Ziegler, Ryan Lowe, Chelsea Voss, Alec Radford, Dario Amodei, and Paul~F Christiano. 2020.
\newblock Learning to summarize with human feedback.
\newblock \emph{Advances in neural information processing systems}, 33:3008--3021.

\bibitem[{Tam et~al.(2024)Tam, Wu, Tsai, Lin, Lee, and Chen}]{tam2024let}
Zhi~Rui Tam, Cheng-Kuang Wu, Yi-Lin Tsai, Chieh-Yen Lin, Hung-yi Lee, and Yun-Nung Chen. 2024.
\newblock Let me speak freely? a study on the impact of format restrictions on performance of large language models.
\newblock \emph{arXiv preprint arXiv:2408.02442}.

\bibitem[{Team(2024)}]{gemini2024}
Gemini Team. 2024.
\newblock \href {https://blog.google/technology/google-deepmind/google-gemini-ai-update-december-2024/} {Introducing gemini 2.0: our new ai model for the agentic era}.
\newblock Accessed: 2025-08-12.

\bibitem[{Team et~al.(2025{\natexlab{a}})Team, Kamath, Ferret, Pathak, Vieillard, Merhej, Perrin, Matejovicova, Ram{\'e}, Rivi{\`e}re et~al.}]{team2025gemma}
Gemma Team, Aishwarya Kamath, Johan Ferret, Shreya Pathak, Nino Vieillard, Ramona Merhej, Sarah Perrin, Tatiana Matejovicova, Alexandre Ram{\'e}, Morgane Rivi{\`e}re, et~al. 2025{\natexlab{a}}.
\newblock Gemma 3 technical report.
\newblock \emph{arXiv preprint arXiv:2503.19786}.

\bibitem[{Team et~al.(2025{\natexlab{b}})Team, Bai, Bao, Chen, Chen, Chen, Chen, Chen, Chen, Chen et~al.}]{team2025kimi}
Kimi Team, Yifan Bai, Yiping Bao, Guanduo Chen, Jiahao Chen, Ningxin Chen, Ruijue Chen, Yanru Chen, Yuankun Chen, Yutian Chen, et~al. 2025{\natexlab{b}}.
\newblock Kimi k2: Open agentic intelligence.
\newblock \emph{arXiv preprint arXiv:2507.20534}.

\bibitem[{Team(2023)}]{vicuna2023vicuna}
Vicuna Team. 2023.
\newblock \href {https://lmsys.org/blog/2023-03-30-vicuna/} {Vicuna: An open-source chatbot impressing gpt-4 with 90\% chatgpt quality}.
\newblock Accessed: 2025-08-12.

\bibitem[{Wang et~al.(2023{\natexlab{a}})Wang, Liang, Meng, Sun, Shi, Li, Xu, Qu, and Zhou}]{wang2023chatgpt}
Jiaan Wang, Yunlong Liang, Fandong Meng, Zengkui Sun, Haoxiang Shi, Zhixu Li, Jinan Xu, Jianfeng Qu, and Jie Zhou. 2023{\natexlab{a}}.
\newblock Is chatgpt a good nlg evaluator? a preliminary study.
\newblock \emph{arXiv preprint arXiv:2303.04048}.

\bibitem[{Wang et~al.(2023{\natexlab{b}})Wang, Li, Chen, Cai, Zhu, Lin, Cao, Liu, Liu, and Sui}]{wang2023faireval}
Peiyi Wang, Lei Li, Liang Chen, Zefan Cai, Dawei Zhu, Binghuai Lin, Yunbo Cao, Qi~Liu, Tianyu Liu, and Zhifang Sui. 2023{\natexlab{b}}.
\newblock Large language models are not fair evaluators.
\newblock \emph{arXiv preprint arXiv:2305.17926}.

\bibitem[{Wei et~al.(2022)Wei, Wang, Schuurmans, Bosma, Xia, Chi, Le, Zhou et~al.}]{wei2022chain}
Jason Wei, Xuezhi Wang, Dale Schuurmans, Maarten Bosma, Fei Xia, Ed~Chi, Quoc~V Le, Denny Zhou, et~al. 2022.
\newblock Chain-of-thought prompting elicits reasoning in large language models.
\newblock \emph{Advances in neural information processing systems}, 35:24824--24837.

\bibitem[{Xu et~al.(2024)Xu, Li, Tao, Shen, Cheng, Li, Xu, Tao, and Zhou}]{xu2024survey}
Xiaohan Xu, Ming Li, Chongyang Tao, Tao Shen, Reynold Cheng, Jinyang Li, Can Xu, Dacheng Tao, and Tianyi Zhou. 2024.
\newblock A survey on knowledge distillation of large language models.
\newblock \emph{arXiv preprint arXiv:2402.13116}.

\bibitem[{Yang et~al.(2025)Yang, Li, Yang, Zhang, Hui, Zheng, Yu, Gao, Huang, Lv et~al.}]{yang2025qwen3}
An~Yang, Anfeng Li, Baosong Yang, Beichen Zhang, Binyuan Hui, Bo~Zheng, Bowen Yu, Chang Gao, Chengen Huang, Chenxu Lv, et~al. 2025.
\newblock Qwen3 technical report.
\newblock \emph{arXiv preprint arXiv:2505.09388}.

\bibitem[{Yu et~al.(2023)Yu, He, Wu, Dai, and Chen}]{yu2023towards}
Zihan Yu, Liang He, Zhen Wu, Xinyu Dai, and Jiajun Chen. 2023.
\newblock Towards better chain-of-thought prompting strategies: A survey.
\newblock \emph{arXiv preprint arXiv:2310.04959}.

\bibitem[{Zheng et~al.(2023)Zheng, Chiang, Sheng, Zhuang, Wu, Zhuang, Lin, Li, Li, Xing et~al.}]{zheng2023judging}
Lianmin Zheng, Wei-Lin Chiang, Ying Sheng, Siyuan Zhuang, Zhanghao Wu, Yonghao Zhuang, Zi~Lin, Zhuohan Li, Dacheng Li, Eric Xing, et~al. 2023.
\newblock Judging llm-as-a-judge with mt-bench and chatbot arena.
\newblock \emph{Advances in neural information processing systems}, 36:46595--46623.

\end{thebibliography}
